\documentclass[]{article}

\usepackage{amsfonts, amsmath, amssymb}
\usepackage{fancyhdr}
\usepackage{graphicx}
\usepackage[left = 3cm, right = 3cm]{geometry}
\usepackage{float}
\usepackage{hyperref}
\usepackage[nottoc, notlot, notlof]{tocbibind}
\usepackage[utf8]{inputenc}
\usepackage[T1]{fontenc}
\usepackage[english]{babel}
\usepackage{lmodern}
\usepackage[round]{natbib}
\usepackage{subcaption}
\usepackage{float}
\usepackage{verbatim}
\usepackage{authblk}
\usepackage{makecell}

\title{Greenhouse gases emissions: estimating corporate non-reported emissions using interpretable machine learning}

\author[1,2]{J\'er\'emi Assael}
\author[3]{Thibaut Heurtebize}
\author[1]{Laurent Carlier}
\author[3]{François Soup\'e}

\affil[1]{BNP Paribas Corporate \& Institutional Banking, Global Markets Data \& Artificial Intelligence Lab, Paris, France}
\affil[2]{Chair of Quantitative Finance, MICS Laboratory, CentraleSup\'elec, Universit\'e Paris-Saclay, Gif-sur-Yvette, France}
\affil[3]{BNP Paribas Asset Management, Quantitative Research Group, Research Lab, Paris, France}

\date{}

\begin{document}

\maketitle

\begin{abstract}
As of 2022, greenhouse gases (GHG) emissions reporting and auditing are not yet compulsory for all companies and methodologies of measurement and estimation are not unified. We propose a machine learning-based model to estimate scope 1 and scope 2 GHG emissions of companies not reporting them yet. Our model, specifically designed to be transparent and completely adapted to this use case, is able to estimate emissions for a large universe of companies. It shows good out-of-sample global performances as well as good out-of-sample granular performances when evaluating it by sectors, by countries or by revenues buckets. We also compare our results to those of other providers and find our estimates to be more accurate. Thanks to the proposed explainability tools using Shapley values, our model is fully interpretable, the user being able to understand which factors split explain the GHG emissions for each particular company.
\end{abstract}



\section{Introduction}  
\label{sec:introduction} 

Past human-being activities or "footprint" are now commonly held responsible for the current pollution of our environment. Our footprint is measures by how fast we consume resources and generate waste versus how fast our planet can absorb our waste and generate resources, according to \cite{wackernagel1998our}. When it comes to our air emissions footprint, the greenhouse gases (GHG) emissions are the most widely analyzed as they allow the calculation of radiative forcing. When this radiative forcing is positive, the Earth system captures more energy than it radiates to space: it is a common measure for the global warming of our planet \citep{hansen2005earth}. The calculation of this carbon footprint tends to account for all GHG emissions caused by an individual, event, organization, service, place or product, and is expressed in units of carbon dioxide equivalent (CO\textsubscript{2}-eq). \\

The annual meetings of the United Nations Climate Change Conference at the World Conferences of the Parties (COP) allow to review the objectives of the global effort to fight climate change. They assess GHG footprints at global level and gather engagement of countries to limit CO\textsubscript{2} emissions for fighting global warming and its impact on biodiversity. In line with these engagements, new definitions, laws and methodologies for calculating and limiting these GHG emissions are voted at country level creating a new framework applicable to companies, the underlying hypothesis being that the country emissions are the sum of emissions coming from its inhabitants and its companies. 

As such, listed and unlisted companies started reporting their emissions in their extra-financial communication. According to \cite{wiedmann2009companies}, the carbon footprint of a company depends on the total amount of CO\textsubscript{2}-eq that is directly and indirectly caused or accumulated over the life stages of its products.  From the companies point of view, the assessment of their GHG footprint can be useful not only for regulatory or accounting disclosure, but also to implement strategies designed to mitigate and reduce their emissions. All frameworks like carbon pricing policies, measuring alignment to climate scenario with the Paris Agreement Capital Transition Assessment (PACTA) or engaging toward net zero GHG emissions via Net Zero Banking Alliances (NZBA) need correct GHG emissions baseline. This momentum will be emphasis by the new Corporate Sustainability Reporting Directive (CSRD) coming into force from 2024 for largest companies to 2026 for Small and Medium-sized Enterprises (SME), in the European Union (EU). This directive will also apply to non-European companies, making over 150 million of euros of turnover in Europe, according to the Council of the European Union. Companies will need to report audited GHG emissions as well as a quantitative pathway and remediation plan to cancel their net emissions. 

Overall, these GHG emissions assessments measure exposure to transition risk and negative cash flows coming from fines or outflows to competitors with greener footprints. They are useful for financial fundamental analysis and slowly implemented in corporate valuation methodologies at least for most vulnerable sectors. Nevertheless as soon as financial institutions aggregate GHG emissions at portfolio level for several companies, they need homogeneous methodologies. At this stage, company reporting of GHG emissions are either voluntary or mandatory depending on location and link to defined nomenclatures (mostly activity types and size of companies). As explained previously, the calculation methodology is often defined along with the regulation and specified at sector level. The heterogeneity of these methodologies can sometimes make comparisons among companies in different countries or sectors difficult and thus create biases. Moreover, not only may calculation methodologies vary but they are also mainly not documented in the reports. \\

In the Global Warming Potential (GWP) framework, for any gas, CO\textsubscript{2}-eq is calculated as the mass of CO\textsubscript{2} which would warm the earth as much as the mass of that gas: it provides a common scale for measuring the climate effects and global warming impacts of different gases. In this way, for a gas with a GWP of 10, two tons of it would have CO\textsubscript{2}-eq of 20 tons. In practice, measuring GHG emissions of a stakeholder requires much more information depending on how the GWP is released. To standardize this methodologies of calculation, the GHG Protocol, first published in 2001 \citep{ghgprotocol2015}, is used by large companies, by the World Business Council for Sustainable Development (WBCSD) and the World Resources Institute (WRI). Even if in some case, companies report according to the ISO 14064 standards or the carbon-balance tool used in France, it has become the most widely used methodology in the world when it comes to assessing GHG emissions. The carbon inventory is divided into three scopes corresponding to direct and indirect emissions:

\begin{itemize}
	\item Scope 1: Sum of direct GHG emissions from sources that are owned or controlled by the company: stationary combustion, e.g. burning oil, gas, coal and others in boilers or furnaces; mobile combustion, e.g. from fuel-burning cars, vans or trucks owned or controlled by the firm; process emissions, e.g. from chemical production in owned or controlled process equipment such as the emissions of CO\textsubscript{2} during cement manufacturing; fugitive emissions from leaks of GHG gases, e.g. from refrigeration or air conditioning units.
	\item Scope 2: Sum of indirect GHG emissions associated with the generation of purchased electricity, steam, heat or cooling consumed by the company. 
	\item Scope 3: Sum of all other indirect emissions that occur in the value chain of the company, including financed emissions via investments. 
\end{itemize}

Most current regulatory standards makes reporting on scope 1 and scope 2 mandatory for large companies. Reporting on scope 3 is mostly optional or to be reported later in 2023 or 2024, even if scope 3, also referred to as value chain emissions, is often the largest component of companies total GHG emissions for some business industries like automakers or financial institutions. In practice, making the methods for calculating emissions in a given industry converge makes it easier not only to model but also to compare the emissions of each company with those of its peers. \\ 

To guarantee data quality of companies reported GHG emissions, independent bodies such as the Carbon Disclosure Project (CDP), a not-for-profit charity that runs the global disclosure system or external auditors in extra financial Corporate Social Responsibility (CSR) reports are more and more involved increasing convergence of methodologies and controls. 

In our study, we limit the methodology to scope 1 and scope 2. Regarding scope 3 emissions, some framework like the Partnership for Carbon Accounting Financials (PCAF), officially recognized by the GHG protocol, allows to measure scope 1 and scope 2 emissions of a financial institution using reported emissions of investments sources but also estimates of scope 1 and 2, as stated in \cite{pcaf2022}. \\

Overall GHG emissions from large firms in developed countries follow a common methodology for calculating scope 1 and 2 emissions: results are either published, validated, or both by independent bodies such as external auditors, the CDP or both. In 2021, this was the case for more than 4 000 companies worldwide, as we observed in this study. For a typical investment universe of 15 000 companies, this means that about 11 000 companies (73\%) were not reporting their scope 1 and 2 GHG emissions. This breadth of reporting is not sustainable even in the short-term, knowing the increasing number of regulatory bodies and investors who either want or are requiring to take into account the GHG emissions of companies. At the same time, even some recent study like \cite{bolton2021global} analyses GHG emissions of 14 468 companies, including 98\% of publicly listed companies, without mentioning or analyzing that 80\% of the data used is coming from GHG modeled estimates from the data provider Trucost. They even construct a regression model to fit all the scopes 1, 2 and 3 data and conclude on global carbon premiums in the market. On the opposite, some studies using the same Trucost dataset specify and analyze more deeply the underlying quality of the GHG emissions data used, like \citep{aswani2022carbon}.  

That is why it is so important to analyze in details the corporate GHG emissions data: operational scopes (accounting consolidation scope, some of biggest factories), standards of calculations (GHG protocol or others), calculation basis (scope 2 Market based versus Location based) and even if the data is modeled (simple derivation from a previous year to more complex non-linear models). When it comes to comparing corporate across geographies and sector or drawing conclusion at global level for anthropogenic GHG emissions we need fair assessment of the GHG emissions at countries, corporations, factories and personal levels. This study is narrowing to unreported estimated emissions of companies. Our model framework focus on estimating the targeted high quality reported emissions for every use cases, especially waiting for international regulatory bodies to bring a homogeneous framework for corporation to report their GHG emissions in extra financial statements.

\section{Literature review - Hypothesis}

Focusing on scopes 1 and 2, available reported data is typically issued from voluntary reporting based on the CDP or on extra financial reports (CSR reports) from companies. With a few exceptions, like France with the Article 173 of the French Energy Transition law, GHG emissions reporting is not yet mandatory but the corporate regulatory framework to report GHG emissions was improve recently with the CSRD in the EU or the Securities and Exchange Commission (SEC) proposed rules in the United States \citep{sec2022}.\\

Corporate GHG emissions models make the link between the industrial processes of each business model and the carbon emissions associated with each stage of those processes. The Environmental Input Output Analysis (EIO), the Process Analysis (PA) models give precise results for a given industrial process \citep{wiedmann2009carbon}. However, neither the information required to quantify companies use of those processes, nor their intensity in the overall annual production chain, is publicly available. Linking detailed industrial processes and technologies with accounting of GHG emission is a perilous task even when it is handle by big corporate sustainable expert teams or by CDP experts.

To mitigate such lack of data, financial data vendors rely on relatively simple models to estimate GHG emissions for some companies that do not currently report. These estimates are usually sector level extrapolations based on indicators such as the number of employees and income generated, or both. Sector averages or regression models constructed from the existing reported GHG emissions data from peer companies have the advantage of simplicity for explainability but the number of regressors is usually limited, as are the sample sizes. Model validation tends to rely on the quality of the regression in-samples where data is available. 

Data providers such as Bloomberg \citep{bbg2022}, MSCI ESG \citep{msci2016, andersson2016hedging, de2016weathered}, Refinitiv ESG – previously known as Thomson Reuters ESG – \citep{refinitiv2017, paribas2016stress, boermans2017pension} and S\&P Global Trucost and CDP use models to estimate the GHG emissions of companies that fail to publish emissions data. Such models relies mainly on rules of proportionality between emissions and the size of the company operations or more recently on more complex approach using non-linear models. The simple models tend to use historical data available for the industry as a basis for the calculation, and focus on predicting the logarithm of GHG emissions. Occasionally, they also use energy specific metric like GHG intensity per the company’s energy consumption and production or per tons of produce cements. However, these metrics are only available for the limited numbers of companies reporting them without reporting their GHG emissions. These models are calibrated on the samples of reported data. Performance is around 60\% in terms of $R^2$ for most samples, when evaluating on the logarithm of the emissions. To be noted, these performance levels are tested in-the-sample, meaning the $R^2$ computed with the logarithm of the GHG emissions is tested with the data used to calibrate the model. The performance of the model is calculated on the same companies used for calculation of the regressions levels. On the other hand, out-of-sample performance test requires a completely new dataset to test the model on unseen companies with reported emissions. Good performance out-of-sample shows that the model avoided overfitting and is able to generalize well. \\

Some more advanced models described in \cite{goldhammer2017estimating, griffin2017relevance, cdp2020methodo} proposed the use of Ordinary Least Squares (OLS) and Gamma Generalized Linear Regression (GGLR) with a broader dataset of publicly available company data for the construction of models. Such models go beyond using just simple factors but relies more usually on data correction processes or smaller sub-samples of industries where the models work correctly. These models are more effective than the previous ones, with in-the-sample $R^2$ computed with the logarithm of the GHG emissions around 80\%. \\

More recently, two studies proposed the use of statistical learning techniques to develop models for predicting corporate GHG emissions from publicly available data. These machine learning approaches take the form of:

\begin{itemize}
	\item in \cite{nguyen2021predicting}, a meta-learner relying on the optimal set of predictors combining OLS, Ridge regression, Lasso regression, ElasticNet, multilayer perceptron, K-nearest neighbors, random forest and extreme gradient boosting as base learners. Their approach generates more accurate predictions than previous models even in out-of-sample situations, i.e., when used to predict reported emissions that were not used to construct the model. Nevertheless, the strongest predictive efficacy of the model was found for predicting aggregate direct and indirect emission scopes as opposed to predicting each of them separately. Furthermore, despite the improvement over existing approaches, the authors also noted that relatively high prediction errors were still found, even in their best model. Indeed, if we consider the five dirtiest industries representing about 90\% of total scope 1 emissions (Utilities, Materials, Energy, Transportation, Capital Goods), their average in-the-sample $R^2$ computed with the logarithm of the GHG emissions is only 51\%. If we consider the five dirtiest industries accounting for about 70\% of the total emissions in terms of scope 2 (Materials, Energy, Utilities, Capital Goods, Automobiles \& Components), their average in-the-sample $R^2$ computed with the logarithm of the GHG emissions is only 52\%. Moreover, their model fails for Insurance, both for scope 1 and scope 2, with $R^2$ of -378\% and -151\%, respectively.
	\item in \cite{bbg2022}, amortized inference with Gradient Boosted Decision Trees (GBDT) models \citep{friedman2001greedy}, re-calibrated using Conditional Mixture of Gammas and Mean Maximum Discrepancy (MMD)-based patterned dropout for regularization. The model is trained on Environment, Social and Governance (ESG) data, fundamentals, and industry segmentation data. The GBDT allows for non-linear patterns to be found even if not all featured data are available. Moreover, an important debiasing approach compares the features distributions for the reporting companies and non-reporting companies by trying to match missing features between labeled data and unlabeled data using MMD. In this model, the $R^2$ computed directly with the GHG emissions go from 84\% for firms with good disclosures (lots of features available) to 41\% for companies with average or poor features disclosures.
\end{itemize}

Understanding the risks and opportunities arising from the GHG emissions of companies requires good financial and non-financial data. In some countries and for some companies, as long as the GHG emissions reporting and auditing is not compulsory, the only viable alternative is to predict non-reported company emissions relying on estimation models. From the above, we believe that the current state-of-the-art does not yet provide good enough models for the task at hand. In our view, the quality of data made available by the specialized data vendors is not yet sufficient. Understanding the reasons behind this problem and being able to propose alternative approaches that can lead to better models and more accurate predictions of unreported data is thus of great importance. \\

The recently proposed approaches by \cite{nguyen2021predicting} and \cite{bbg2022}, based on statistical learning, offer a promising starting point. The central challenge with such statistical learning approaches is to strike the right balance between increasing both the model complexity and accuracy while limiting the risk of overfitting. In this paper, we propose a statistical learning model to predict unreported scope 1 and scope 2 company emissions in an investment universe of about 50,000 companies of which only about 4,000 companies actually report. This model is inspired by the work of \cite{heurtebize2022corporate}, and aims at achieving the following qualities:

\begin{itemize}
	\item important final coverage, aiming at using the model for smaller and private companies in a scope of 50,000 companies.
	\item simplicity and interpretability, making sure the statistical explanations of the outputs are clear and exhaustive enough.
	\item implementation ease with no manual data correction.
	\item flexibility, making sure the model can be adapted to all types of usages within a set of different calibrations, and to be able to easily include new input data with the evolution of regulations, especially on disclosure.
\end{itemize}

To succeed in this, we made some significant choices in departing from existing approaches. First, our models are always tested on data samples never seen during the calibration, so that we can truly measure their generalization abilities. The second important decision was to keep model complexity to a minimum by relying only on the most accurate non-linear machine learning approaches. The last important decision was to keep the raw dataset from data providers with no manual corrections. Indeed, even if the use of incorrect features or emissions data can reduce the accuracy of models, their industrialization brings some operational constraints to produce automatic updates of the model. Thus, we introduce shortly an automatic data polishing process at the end of this study. 
In the remainder of the paper, we shall describe the data retained to calibrate and evaluate our model. This will be followed by sections in which we present in depth the designed methodology, insisting on our particular implementation choices, necessary to apply it to the use case of estimating GHG scope 1 and 2 emissions. We will then discuss the results associated to this methodology both by comparing our estimates to true reported by companies GHG emissions and by comparing our estimates to the ones from other providers.  Finally, we will provide tools to get insights on how our model works and why it estimates such values of GHG emissions.

\section{Datasets}
\label{sec:datasets} 

We have at our disposal an important variety of data sources. Following \cite{heurtebize2022corporate}, we rely on two sets of indicators. The first set refers to data retrieved at the company level. For a given company, we gather all indicators exhibit in Tab. \ref{tab:comp_indicators}, selecting yearly data. Such indicators allows to get a sense at the company profitability, assets size, assets location and how they are used.

The second set of indicators are regional ones, also selected each year, and presented in Tab. \ref{tab:reg_indicators}. They provide information on the environment the company is incorporated in.	

Data are extracted between 2010 and 2020 from Refinitiv Worldscope database, for a total of 531 408 samples. It represents 65 673 companies between 2010 and 2020, incorporated in 115 countries, with 48 429 companies incorporated in 112 countries for 2020 alone.

\begin{table}
\centering
	
\begin{subtable}{\textwidth}
	\centering
	\resizebox{\textwidth}{!}{\begin{tabular}{l|l|l}
		\textbf{Type of indicator} & \textbf{Data Provider} & \textbf{Name of indicator} \\
		\hline
		\hline
		General & Refinitiv & Country of Incorporation \\
		\hline
		General & Refinitiv & Employees \\
		\hline
		Industry Classification & Bloomberg & BICS Classification Levels 1 to 7 \\
		\hline
		Industry Classification & Bloomberg & New Energy Exposure Rating \\
		\hline
		Financial & Refinitiv & Accumulated Depreciation \\
		\hline
		Financial & Refinitiv & Capital Expenditure \\
		\hline
		Financial & Refinitiv & Depreciation, Depletion \& Amortization \\
		\hline
		Financial & Refinitiv & Enterprise Value \\
		\hline
		Financial & Refinitiv & Revenues \\
		\hline
		Financial & Refinitiv & Property, Plant \& Equipment - Gross \\
		\hline
		Financial & Refinitiv & Property, Plant \& Equipment - Net \\
		\hline
		Financial & Bloomberg & Corporate Actions \\
		\hline
		Energy & Bloomberg & Energy Consumption \\
		\hline
		Energy & Bloomberg & Total Power Generated \\
		\hline
		Greenhouse Gases Emissions & Bloomberg & Reported GHG Emission - Scope 1 \\
		\hline
		Greenhouse Gases Emissions & Bloomberg & Reported GHG Emission - Scope 2 \\
		\hline
		Greenhouse Gases Emissions & Carbon Disclosure Project & Reported GHG Emission - Level 7 quality - Scope 1 \\
		\hline
		Greenhouse Gases Emissions & Carbon Disclosure Project & Reported GHG Emission - Level 7 quality - Scope 2 \\
	\end{tabular}}
	\caption{Indicators retrieved at the company level. BICS refers to the Bloomberg Industry Classification Standard.}
	\label{tab:comp_indicators}
\end{subtable}

\vspace{2em}

\begin{subtable}{\textwidth}
	\centering
	\resizebox{\textwidth}{!}{\begin{tabular}{l|l|l}
		\textbf{Type of indicator} & \textbf{Data Provider} & \textbf{Name of indicator} \\
		\hline
		\hline
		Regional & International Energy Agency & Country Energy Mix Carbon Intensity \\
		\hline
		Regional & WorldBank & Existence of an Emission Trading System\\
		\hline
		Regional & WorldBank & Existence of carbon taxes \\
	\end{tabular}}
	\caption{Indicators retrieved at the regional level for each country or sub-region in which a company is incorporated in.}
	\label{tab:reg_indicators}
\end{subtable}

\caption{Data sources and indicators used in the model.}
\label{tab:indicators}
\end{table}

\section{Methods}

\subsection{Problem settings}
\label{sub:problemsettings}

Our goal is to develop a data-driven model estimating scope 1 and scope 2 greenhouse gases emissions of a company which has not reported them. Using the vast amount of available indicators, whose selected ones have been exhibit in Section \ref{sec:datasets}, we can build a high quality dataset and calibrate a machine learning model which output the estimated emission of a company. This automatized method allows to estimate the emissions of any company as long data related to it can be gathered. Scope 1 and scope 2 emissions will be estimated through two separate models. \\

We are in a regression setting: the model learns for each possible couple $(i,t)$ the reported emission $Y_{i,t}$ from a set of $P$ potentially explanatory factors called features. Here, $i$ represents a company and $t$ the year of sampling of the features and emission. Let us relabel all the couples $(i,t)$ by the index $n\in\{1,\cdots,N\}$. This regression problem consists in explaining $Y_n$, the reported emission, by a vector $X_n$ with $P$ components, or equivalently, to explain the vector $Y \in \mathbb{R}^{N}$ from the lines of matrix $X\in\mathbb{R}^{N \times P}$. $Y$ is called the target and $X$ the features matrix. The problem is then to train a machine learning method to learn the mapping between the lines of $X$ and the components of the vector $Y$. Once the training is complete, such a model takes as input a vector of features and outputs the estimated greenhouse gas emission. \\

The state of the art for regression problems on tabular data like this one is provided by Gradient Boosting models \citep{friedman2001greedy}, as shown for instance in \cite{shwartz2021tabular}. Gradient boosting consists in using a sequence of weak learners, making wrong predictions, that iteratively correct the mistakes of the previous ones, to eventually yields a strong learner, making good predictions. We use here decision trees as weak learners: we are using the GBDT algorithm. Different implementations of the GBDT method has been proposed, e.g. XGBoost \citep{chen2016xgboost}, LightGBM \citep{ke2017lightgbm}, CatBoost \citep{prokhorenkova2017catboost}. We use here LightGBM. The advantage of such methods with respect to linear regression is that they are able to learn more generic functional forms.

The models are trained to minimize the mean-squared error (cost function), also refers in this paper as MSE, defined as
$$
\mathcal{L} = -\frac{1}{N} \sum_{i=1}^{N} | y_i - \hat{y}_i | ^ 2,
$$
where $\hat{y}_i$ is the predicted output from the model (decimal logarithm of the GHG estimation, as explained in section \ref{sec:targcleanproc}) and $y_i$ is the ground truth (decimal logarithm of the reported emission).

\subsection{Target computation}

\subsubsection{Raw target obtention}
\label{sec:raw_target_obtention}

The explained variable, the reported GHG emissions for scopes 1 and 2, are sourced using two databases:
\begin{itemize}
	\item CDP data. We are using the non-modeled and audited emissions from CDP which are at level 7, the highest quality level. Details on CDP methodology and quality review are available in their documentation \citep{cdp2020methodo}.
	\item Bloomberg data. We are using the reported GHG emissions gathered by Bloomberg, sourced from the companies extra-financial communication. 
\end{itemize} 

As the CDP data is reviewed, with direct contact with the considered company or following the company answers to CDP questionnaires, we choose to prioritize the CDP reported emissions when both data sources report a emission for this company. There reported emissions are expressed in tCO\textsubscript{2}-eq.

\subsubsection{Target cleaning procedure}
\label{sec:targcleanproc}

GHG emissions are reported at different dates during the year. To unify samples and keep meaning with the used training features, we attribute the GHG emissions reported between January and June of the year $y$ to the year $y-1$ and the GHG emissions reported between July and December of the year $y$ to the same year $y$. For both scopes, we remain with only one reported GHG emission per company and per year. \\

Variability is an important characteristic of GHG emissions data, leading to sometimes inconsistencies, with important changes in emissions for a company over the years: this could be due to changes in the reporting methodology, to a corporate action like the acquisition of a subsidiary or mergers... The chosen cleaning procedures mitigates these issues: we develop for this purpose a jump cleaning methodology. 

We call \textit{jump} a year-to-year variation in the GHG emission reported value of a company bigger than a threshold of 50\%. This jump processing procedure aims at spotting jumps inside the dataset, removing all inconsistent points unless they can be explained by a significant corporate action. We make the hypothesis that the most recent data is the highest quality one: if an unexplained jump is detected in the time serie of GHG emissions of a company, all data points before the jump and the jump are removed. A jump is unexplained if we cannot find a concomitant and large enough corporate action to justify it. In practice, a jump is said \textit{explained} if we find, using a Bloomberg corporate action dataset, at least one corporate action amounting for at least 20\% of the company revenues during the year before or after the considered jump. The different thresholds were determined by trials and errors. \\

To reduce the negative impact of the skewed nature of the GHG emissions distribution, we are training our model to estimate the decimal logarithm of the GHG emissions instead of the raw value. Another advantage of using the decimal logarithm resides in the interpretation of the estimated value. A error of one unit in the decimal logarithm estimation mean an error of one order of magnitude in the raw GHG emission (multiplied/divided by 10 in comparison to the true value).

\subsection{Training features}

For each of the obtained target, we fetch a vector of features using the data sources exposed in Tab. \ref{tab:indicators}. As two models are trained for the scopes 1 and 2, we obtain two feature matrices, representing the training features for each of the scope. The scope 1 training set has 16234 samples and the scope 2 16925. We summarize in Tab. \ref{tab:cat_features_summary} and \ref{tab:num_features_summary} the 21 features used to train the model as well as their distribution and average coverage in the two training sets. In the remainder of this section, we provide details on these different features. Missing values are let as such as the LightGBM model is able to handle them while preserving performance. 

\begin{table}
	\centering
	\resizebox{\textwidth}{!}{\begin{tabular}{l|l|l|l}
		\textbf{Type of feature} & \textbf{Name} & \textbf{Values} & \textbf{Coverage} \\
		\hline
		\hline
		General  & Year & 2010 to 2020 & 100\% \\
		\hline
		General  & Country of Incorporation & \makecell[l]{Country code \\ (ISO 3166, alpha-3 code)} & 100\% \\
		\hline
		Industry Classification & BICS Classification Levels 1 to 7 & Industry Name & 100\% \\
		\hline
		Industry Classification & New Energy Exposure Rating &  \makecell[l]{A1 Main driver: 50 to 100\% \\ A2 Considerable: 25 to 49\% \\ A3 Moderate: 10 to 24\% \\ A4 Minor: less than 10\% \\ NaN} & 54.1\% \\
		\hline
		Regional & \makecell[l]{CO\textsubscript{2} Law: \\ Existence of an ETS or carbon taxes} & \makecell[l]{National Implemented \\ Subnational Implemented \\ No CO\textsubscript{2} Law} & 100\% \\
	\end{tabular}}
	\caption{Categorical features used to train the GHG emissions estimation model.}
	\label{tab:cat_features_summary}
\end{table}

\begin{table}
	\centering
	\resizebox{\textwidth}{!}{\begin{tabular}{l|l|l|l|l|l|l}
		\textbf{Type of feature} & \textbf{Name} & \makecell[l]{\textbf{1st} \\ \textbf{percentile}} & \textbf{Median} & \makecell[l]{\textbf{99th} \\ \textbf{percentile}} & \textbf{Unit} & \textbf{Coverage} \\
		\hline
		\hline
		General & Employees & 73 & 11 810 & 330 000 & / & 87.3\% \\
		\hline
		Financial & Capital Expenditure & 0 & 204 & 118 374 & Million \$ & 99.8\% \\
		\hline
		Financial & Enterprise Value & 11.4 &  7 578 & 2 609 476 & Million \$ & 99.5\% \\
		\hline
		Financial & Revenues & 56.3 &  4 167 & 1 939 292 & Million \$ & 100\% \\
		\hline
		Financial & \makecell[l]{Property, Plant \\ \& Equipment Gross} & 28.6 & 3 291 & 1 896 412 & Million \$ & 87.2\% \\
		\hline
		Financial & \makecell[l]{Property, Plant \\ \& Equipment Net} & 8.4 & 1 542 & 966 459 & Million \$ & 99.6\% \\
		\hline
		Financial & Life Expectancy of Assets & 0.42 & 13.42 & 50 & Year & 99.2\% \\
		\hline
		Energy & Energy Consumption & 1.7 & 731 & 207 784 & GWh & 74.1\% \\
		\hline
		Energy & Total Power Generated & 0.1 & 20 900 & 564 436 & GWh & 3.3\% \\
		\hline
		Regional & \makecell[l]{Country Energy Mix \\ Carbon Intensity} & 17.7 & 53.0 & 76.9 & t CO\textsubscript{2}/TJ & 99.8\% \\
	\end{tabular}}
	\caption{Numerical features used to train the GHG emissions estimation model.}
	\label{tab:num_features_summary}
\end{table}

\subsubsection{Financial features}

The model relies on financial features, allowing a better understanding of the size of a company and its assets. The Capital Expenditure, Enterprise Value, Gross Property Plant \& Equipment (GPPE), Net Property Plant \& Equipment (NPPE) and Revenues features are obtained annually for each companies for which we have a target, meaning a reported GHG emission for scope 1 and/or scope 2. These values are converted from the reporting currency to dollars using the foreign exchange rate from the 31\textsuperscript{st} December of the considered year.

The last financial feature, the Life Expectancy of Assets, is obtained following \cite{griffin2017relevance} and \cite{nguyen2021predicting} using the following formula:

$$
Life Expectancy  = \frac{GPPE}{Depreciation Expense}
$$

The idea behind this proxy is to estimate the average life expectancy of the assets of a company by dividing the total amount of tangible assets of a company by the depreciation expense the company reported for the considered year. We make the hypothesis that a company whose assets have a longer life expectancy are in average older and may emit more GHG.

As we do not have the Depreciation Expense indicator in our dataset and the Gross property Plant \& Equipment feature have a lot of missing values, we use the equivalent following formula:

$$
Life Expectancy  = \frac{NPPE - Capital Expenditure + Accumulated Depreciation}{Depreciation, Depletion \& Amortization}
$$

The numerator is modified by decomposing the Gross Property Plant \& Equipment term. We make the approximation that, if the Capital Expenditure or Accumulated Depreciation indicators are missing values, they are ignored and their values are set to 0. The denominator is modified by adding the depletion and amortization expense. We did not measure a significant impact of these approximations on the final GHG emission estimation.

\subsubsection{Industry classification}

Industry classification features allows the model to grasp the nature of the assets of a company by understanding in which fields they are used. The GHG emission profile of companies operating in different sectors are different. There exists numerous industry classifications, grouping companies differently: this feature is critical for our model, as we do not want to rely on a classification which would never, at any level, make the difference between companies operating in the Oil \& Gas, Renewable Energy or Nuclear fields. We rely on the industry classification feature which gave us the best results and is detailed enough so that the model is about to grasp granular differences in the companies profiles. 

For each company, we get its main industry classification using the Bloomberg Industry Classification Standard (BICS). The industry in which a company is classified corresponds to the one in which it is making the biggest faction of its revenues. The BICS classification is a detailed one, with a good level of granularity with seven hierarchical levels. It makes granular distinctions between the different sector, going as far as distinguishing companies both operating in the Oil \& Gas Production field but either working on Petroleum Marketing or focusing on Exploration \& Production.

With the important level of details of the BICS classification, we do not have enough density for the deeper levels in our training dataset: not all companies have a data for levels 5, 6 or 7 in the classification. As a result, having just a few instances of a particular industry at a deep level is only adding noise to the model and make it more prompt to overfitting, the model having more difficulties to generalize to other samples. In our preprocessing, we remove all occurrences of industries that are present less than 10 times in the training set. They are replaced by a NaN value, missing values being directly handled by the LightGBM model. \\

As precise as the BICS classification is, we complement it with the New Energy Exposure Rating from Bloomberg. It is a categorical feature which estimates the percentage of an organization's value that is attributable to its activities in renewable energy, energy smart technologies, Carbon Capture and Storage (CCS) and carbon markets. This categorical data can take five values:
\begin{itemize}
	\item A1 Main driver: 50 to 100\% of the organization's value is estimated to derive from these activities.
	\item A2 Considerable: 25 to 49\% of the organization's value is estimated to derive from these activities.
	\item A3 Moderate: 10 to 24\% of the organization's value is estimated to derive from these activities.
	\item A4 Minor: less than 10\% of the organization's value is estimated to derive from these activities.
	\item NaN if missing.
\end{itemize}

\subsubsection{Energy data}

Energy features, expressed in GWh, are often directly correlated to the GHG emissions and allows the model to have a better understanding of how a company is using its assets. Energy Consumption is the amount of energy consumed by a company during a year. Total Power Generated can be explained as the energy produced in a year by a company and therefore, it is only relevant for companies in some specific industries, explaining the low coverage shown in Tab. \ref{tab:num_features_summary}. To be noted the distinction between renewable and non-renewable power generated is available in our dataset but was not used in this version of the model. The reporting period may differ between companies: similarly to the GHG emissions targets, we attribute the values reported between January and June of the year $y$ to the year $y-1$ and those reported between July and December of the year $y$ to the same year $y$. 

\subsubsection{Regional data}

Regional data allows the model to get a sense of the environment the company is operating in, for the country in which it is incorporated in. The Carbon Intensity of Energy Mix refers to the CO\textsubscript{2} Emissions from fuel combustion for the country in which the considered company is incorporated in. Data are gathered from the International Energy Agency (IEA). Depending on when these data are obtained, there may be missing data for the most recent years: in this case, the time serie for the considered country is extended using the last known value.

The model also relies on a categorical feature describing if a system of carbon taxes or an Emission Trading System (ETS) have been put in place at a national or sub-national level. This feature, called CO\textsubscript{2} Law, can take three values:
\begin{itemize}
	\item No CO\textsubscript{2} law: no carbon tax or ETS has been put in place for the considered country;
	\item National Implemented: one or both of these systems are implemented in the whole considered country.
	\item Sub-national Implemented: one or both of these systems are implemented in part of the considered country (a state in Canada on in the USA for instance).
\end{itemize}

\subsection{High Quality Dataset}

\begin{figure}
	\centering
	\includegraphics[width=0.5\linewidth]{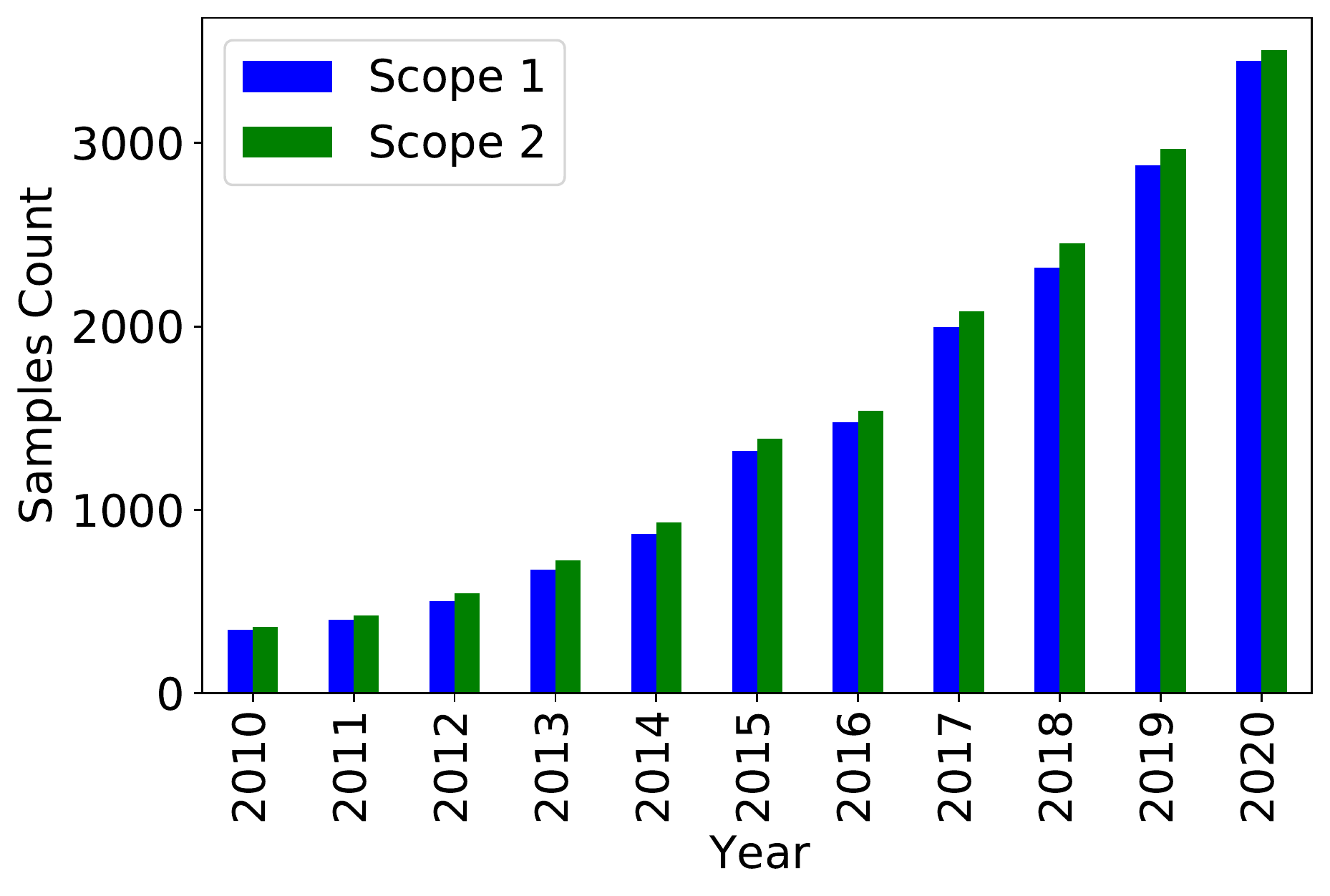}
	\caption{GHG emissions: number of companies with a reported emission per year for scopes 1 and 2}
	\label{fig:samples_count_ghg}
\end{figure}

Using the features and target cleaning procedures, we obtain the final training datasets to estimate scope 1 and scope 2 GHG emissions. These two high quality datasets are used in all the remaining parts of this study. Figure \ref{fig:samples_count_ghg} shows for scope 1 and scope 2 the number of companies for which a reported GHG emission per year was obtained. We observe an important increase of data quantity through the years, which illustrates the growing importance of GHG emissions reporting.

\subsection{Cross-validation and hyperparameter tuning -- Out-of-sample performance evaluation}
\label{sub:splittingstrat}

The usual strategy in machine learning for time series consists of a single data split into causal consecutive train, validation and test data sets. The model learns the mapping between features and target on the training set, determined its parameters on the validation one and is finally tested on the test one. To avoid overfitting and preserve the generalization capacity of the model, the test set should only be used at the end of the training, to evaluate the model. This usual strategy is not appropriate for the current problem, estimating the GHG emissions of companies which has not reported them during the last year. Indeed: 
\begin{itemize}
	\item the usual splitting scheme does not comply with the use case: the goal is not to predict future GHG emissions but to estimate unreported ones during the last available year. 
	\item the amount of data grows from a very low baseline both quantity- and quality-wise. Oldest data are not exploitable alone: using this splitting scheme would lead to unreliable results as only old data would be in the training set. We want to rely on the entire time span of data we have.
	\item similarly, GHG emissions data are non-stationary, leading to inaccurate results using this standard splitting scheme.
\end{itemize}

To address these issues, we developed a specific testing methodology and cross-validation scheme, inspired by \cite{assael2022esg}. \\

As we want to estimate unreported data during the last available year, the test set built to evaluate our models should only include companies which are not in the training or validation sets: we want to estimate unreported emissions of companies which, most of the time, never reported their emissions before. Moreover, it allows us to avoid a potential bias: because of the huge year-over-year correlation of GHG emissions for a same company, having the same company both in training/validation and in test during different years would lead to an overfitted model. In practice, we build the test set by selecting 30\% of the companies for which we have a reported value during the last available year: these samples constitute the test set. These companies may have other reported emissions for other years: all these companies are removed from the training and validation sets. 

For training and validation, we use a $K$-fold company-wise cross-validation: 80\% of companies are randomly assigned to the training set and the remaining 20\% to the validation one. We train 180 models on each of the $K$ training sets varying the hyperparameters of the LightGBM algorithm and select the best one based on its average performances, measured using the MSE, on the respective validation sets. In this way, we respect the framework we are in, not having any company both in training and in validation and models are trained with a large part of the most recent and more relevant data, while validating the model also with the most recent and and more relevant data. We take $K=4$.  

\begin{figure}
	\centering
	\includegraphics[width=1\linewidth]{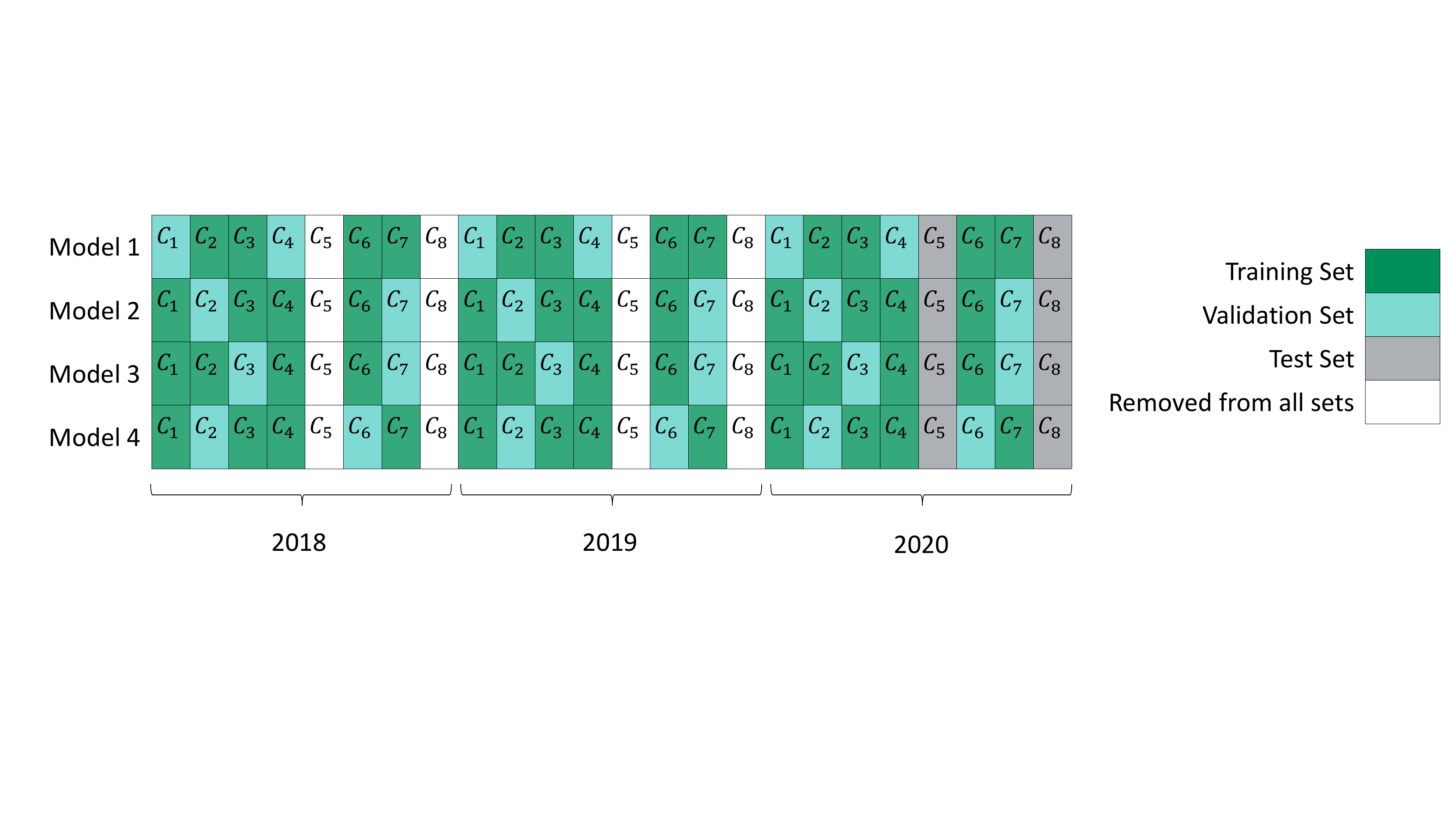}  
	\caption{Company-wise cross-validation: the validation sets consists of randomly selected companies, which allows training to account for most of the most recent data.}
	\label{fig:cross_val_scheme}
\end{figure}

Figure \ref{fig:cross_val_scheme} illustrates on a three years and eight companies dataset the used procedure to build our training, validation and test sets.

\section{Results: evaluating the performances of the model}
\label{sec:results}

We first assess the quality and performances of the model on the designed high-quality testing set, build as explained in section \ref{sub:splittingstrat}. 

\paragraph{Selected metrics}

An important contribution of this work is to design a model with both good global performances on the test set and good performances for each business sector, at different levels of granularity, for each country and each decile of revenues. 

To evaluate performances on the test set, the selected metric is the root mean-squared error also refers in this paper as RMSE, defined as
$$
RMSE = \sqrt{\frac{1}{N} \sum_{i=1}^{N} || y_i - \hat{y}_i ||^2}
$$
where $\hat{y}_i$ is the GHG estimation (log-transformed) from the model and $y_i$ is the ground truth (log-transformed). \\

Another measure of performance is the mean-absolute error, refers in this paper as MAE, defined as
$$
MAE = \frac{1}{N} \sum_{i=1}^{N} | y_i - \hat{y}_i |
$$
where $\hat{y}_i$ is the GHG estimation (log-transformed) from the model and $y_i$ is the ground truth (log-transformed). \\

The $R^2$ metric is also a common measure of performance. It is defined as:
$$
R^2 = 1 - -\frac{\frac{1}{N} \sum_{i=1}^{N} (y_i - \hat{y}_i)^2}{\frac{1}{N} \sum_{i=1}^{N}  (y_i - \bar{y}_i)^2}
$$
where $\hat{y}_i$ is the GHG estimation (log-transformed) from the model, $y_i$ is the ground truth (log-transformed) and $\bar{y}_i$ is the average of the ground truth emissions (log-transformed). \\

We provide global results using these three different metrics for comparability purposes across the literature on GHG emissions models. RMSE and MAE metrics can vary between 0 and infinity, a value of 0 meaning that the model is perfectly accurate. Regarding the $R^2$ metric, it varies between 0 and 1, a value of 1 meaning that the model is perfectly accurate.
RMSE and MAE are easier to interpret than $R^2$ in the context of GHG emissions as they are expressed in the same unit as the log-transformed GHG emission. RMSE penalizes more than MAE larger errors: large errors are undesirable in the context of estimating GHG emissions, that's why we rely mainly on the RMSE metric in the remainder of this study.

\paragraph{Multiple test sets}

As shown in Fig. \ref{fig:samples_count_ghg}, we do not have a great quantity of samples to train our model: this leads to small test sets with around 800 data points. As a result, evaluation on the test set may be subject to a high variability: a few single wrongly estimated point could lead to an important deterioration of performance. We mitigate this issue by creating five different test sets and evaluate the model performances on these five test sets.

\subsection{Global performances}

Table \ref{tab:global_metrics_results} displays the mean global results of the scope 1 and scope 2 models for the RMSE, MAE and $R^2$ metrics on each test sets. In comparison to the literature like \cite{goldhammer2017estimating} or \cite{griffin2017relevance}, we display only out-of-sample results as they are the ones that really show the performance of the model and its capacity to generalize beyond the training data. 

These metrics are computed using the decimal logarithm of the predicted emission and the decimal logarithm of the reported emission. It means for instance that a RMSE of 1 corresponds to an error of one order of magnitude in terms of raw greenhouse gas emissions. As stated as an introduction to this section, RMSE, MAE and $R^2$ metrics are displayed for comparability purposes across the literature. As they differ in their definition, they should not be compared to each other.

\begin{table}
	\centering
	\begin{tabular}{l l||c|c|c|c}
		& & \multicolumn{2}{c|}{\textbf{Scope 1}} & \multicolumn{2}{c}{\textbf{Scope 2}} \\
		\hline
		\textbf{Range} &  \textbf{Metric} & \textbf{Mean} & \textbf{Standard Deviation} & \textbf{Mean} & \textbf{Standard Deviation}\\
		\hline
		\hline
		$[0, 1]$ & $R^2$ & 0.832 & 0.007 & 0.746 & 0.017 \\
		\hline
		\hline
		$[0, +\inf[$ & RMSE & 0.578 & 0.007  & 0.522 & 0.031 \\
		\hline
		\hline
		$[0, +\inf[$ & MAE & 0.401 & 0.006 & 0.341 & 0.010 \\
	\end{tabular}
	\caption{Results of the model on the five different test sets: mean and standard deviation of the $R^2$, RMSE and MAE metrics. The three metrics, computed on the decimal logarithm of the emissions, are given for comparability purposes across the literature and should not be compared to each other}
	\label{tab:global_metrics_results}
\end{table}

\subsection{Breakdown of performances by sectors, countries and revenues}

Beside assessing the global performance of the models, we consider a breakdown of the models performances per sectors, per countries and per revenues: it allows for a transparent review of the performances of the model and to understand better its strengths and weaknesses. \\

\begin{figure}
	\centering
	\begin{subfigure}{0.80\textwidth}
		\centering
		\includegraphics[width=1\textwidth]{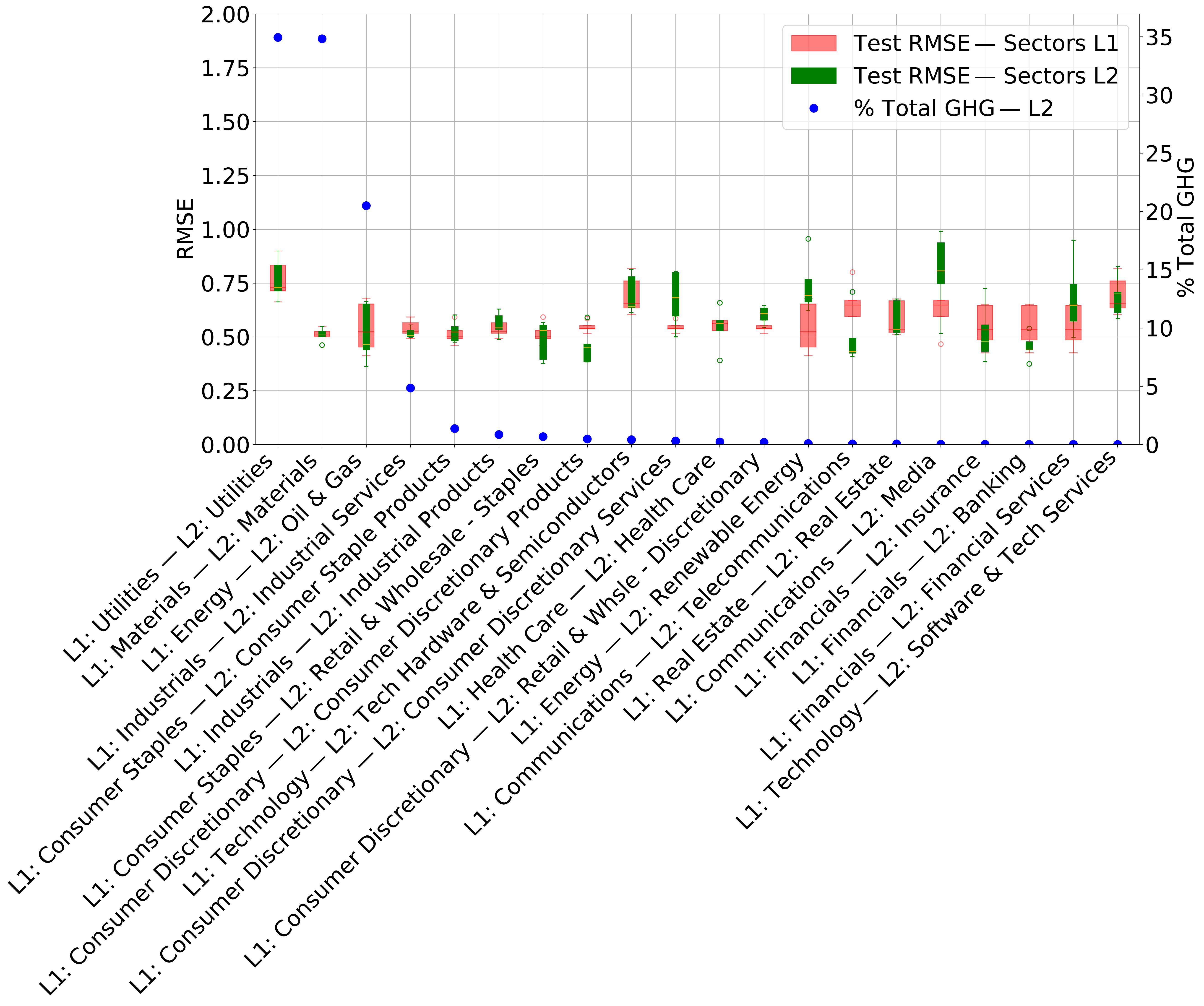}
		\caption{Boxplot of test RMSE on the five different test sets per BICS sectors levels 1 and 2, ordered by level 2 sectors emissions.}
		\label{fig:most_emissive_lvl12_RMSE_CF1_2020_LGBM_BICS_BICS_TPS_raw_v16}
	\end{subfigure}
	\hfill
	\begin{subfigure}{0.49\textwidth}
		\centering
		\includegraphics[width=1\textwidth]{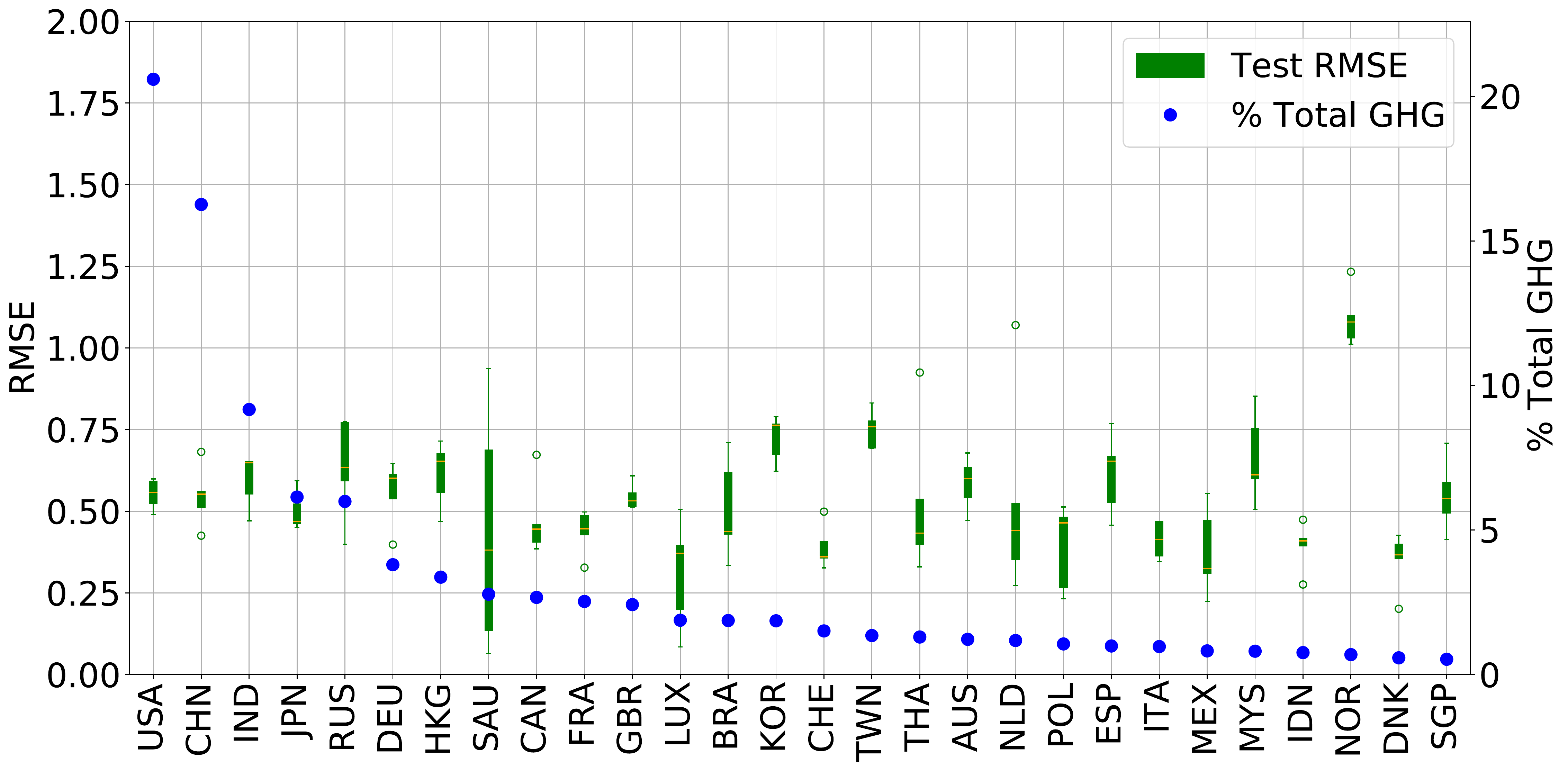}
		\caption{Boxplot of test RMSE on the five different test sets per countries, ordered by countries emissions.}
		\label{fig:most_emissive_RMSE_CF1_2020_LGBM_BICS_IsoCtry_TPS_raw_v16_False}
	\end{subfigure}
	\hfill
	\begin{subfigure}{0.49\textwidth}
		\centering
		\includegraphics[width=1\textwidth]{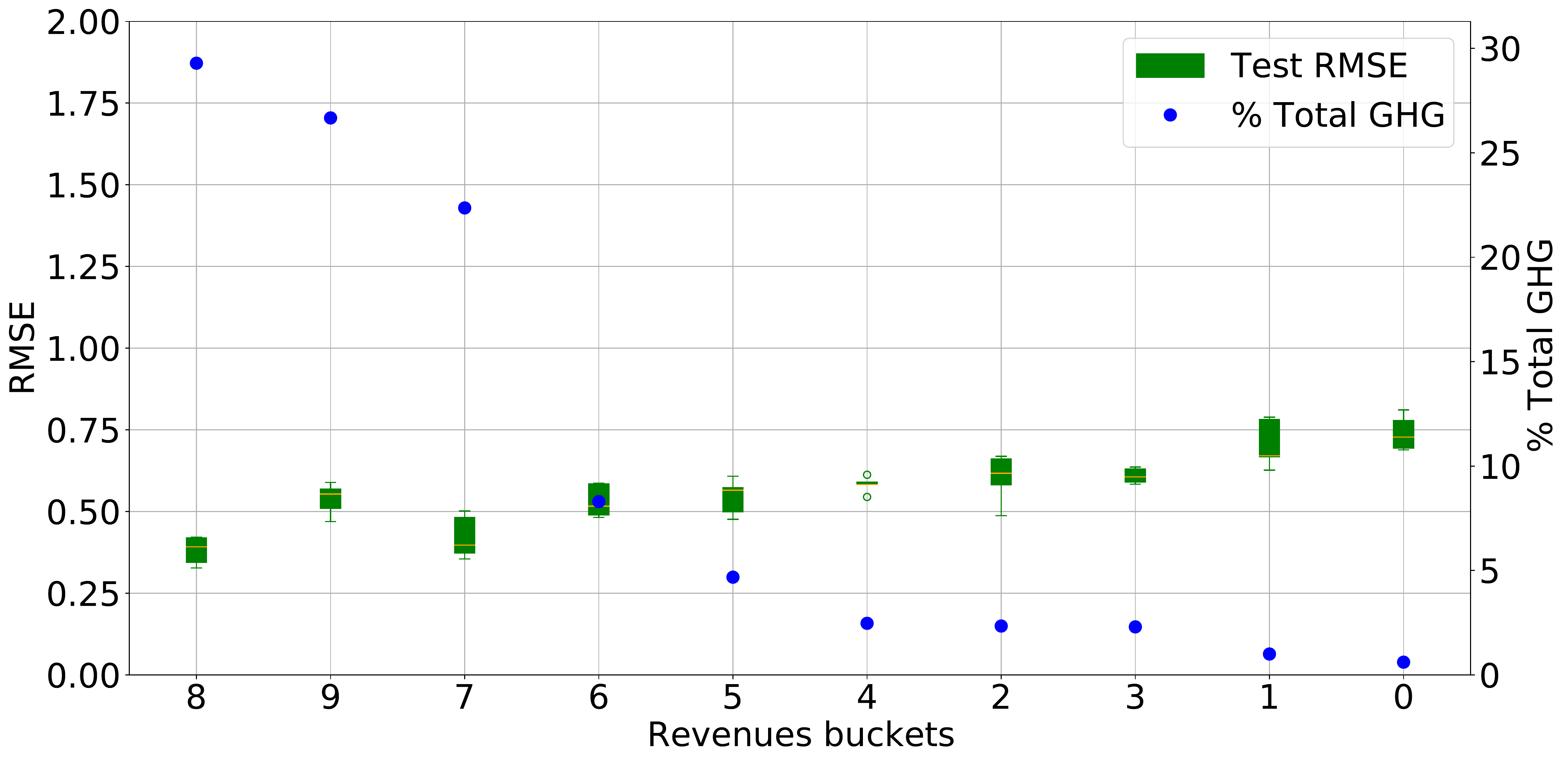}
		\caption{Boxplot of test RMSE on the five different test sets per deciles of revenues, ordered by revenues deciles emissions.}
		\label{fig:most_emissive_RMSE_CF1_2020_LGBM_BICS_Revenues Bucket_TPS_raw_v16_False}
	\end{subfigure}

	\caption{GHG emissions scope 1: distribution of performances of the model on five test sets according to different characteristics of companies.}
	\label{fig:ghg_emissions_scope1_summary_perf}
\end{figure}

\begin{figure}
	\centering
	\begin{subfigure}{0.8\textwidth}
		\centering
		\includegraphics[width=1\textwidth]{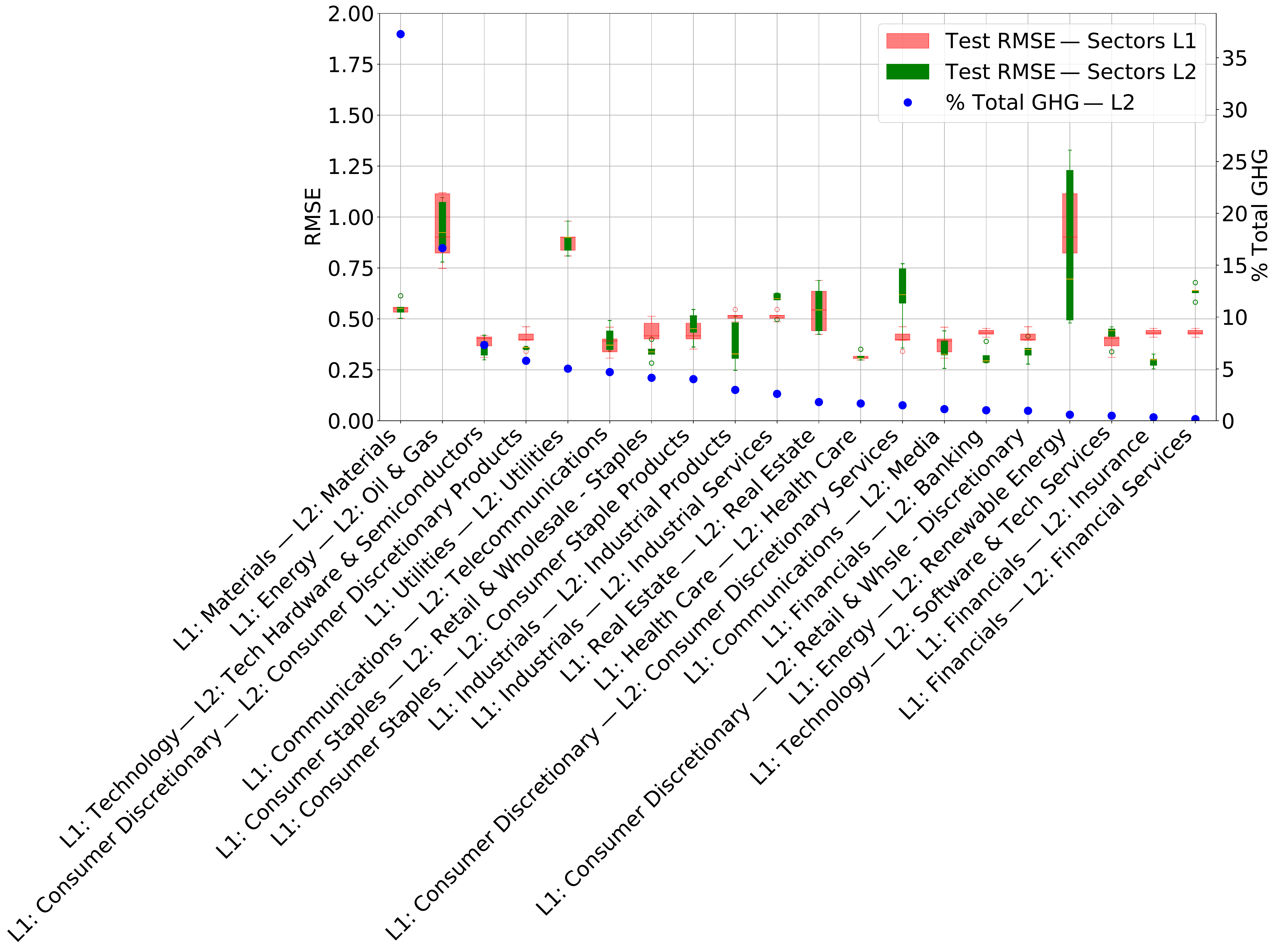}
		\caption{Boxplot of test RMSE on the five different test sets per BICS sectors levels 1 and 2, ordered by level 2 sectors emissions.}
		\label{fig:most_emissive_lvl12_RMSE_CF2_2020_LGBM_BICS_BICS_TPS_raw_v16}
	\end{subfigure}
	\hfill
	\begin{subfigure}{0.49\textwidth}
		\centering
		\includegraphics[width=1\textwidth]{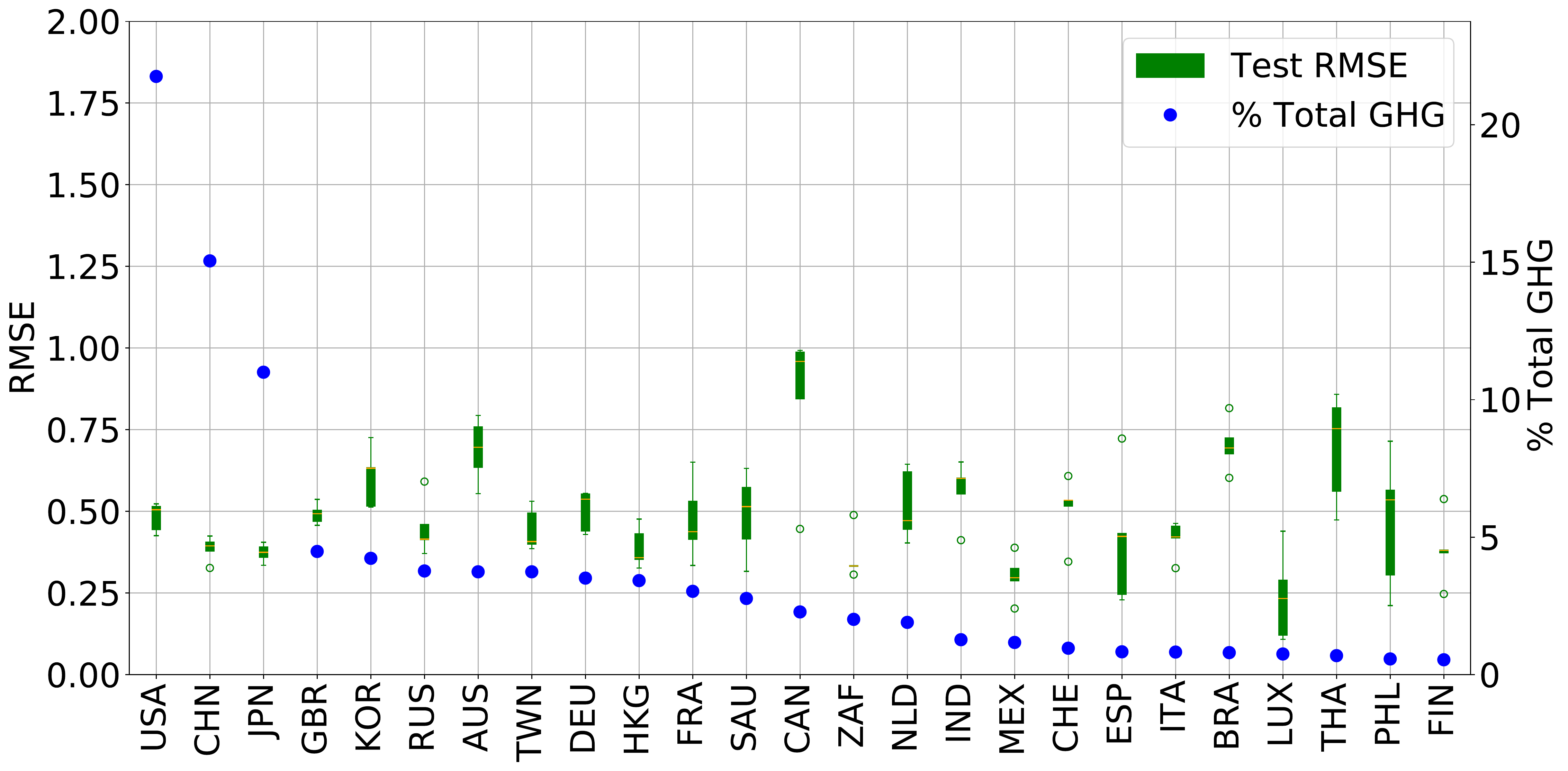}
		\caption{Boxplot of test RMSE on the five different test sets per countries, ordered by countries emissions.}
		\label{fig:most_emissive_RMSE_CF2_2020_LGBM_BICS_IsoCtry_TPS_raw_v16_False}
	\end{subfigure}
	\hfill
	\begin{subfigure}{0.49\textwidth}
		\centering
		\includegraphics[width=1\textwidth]{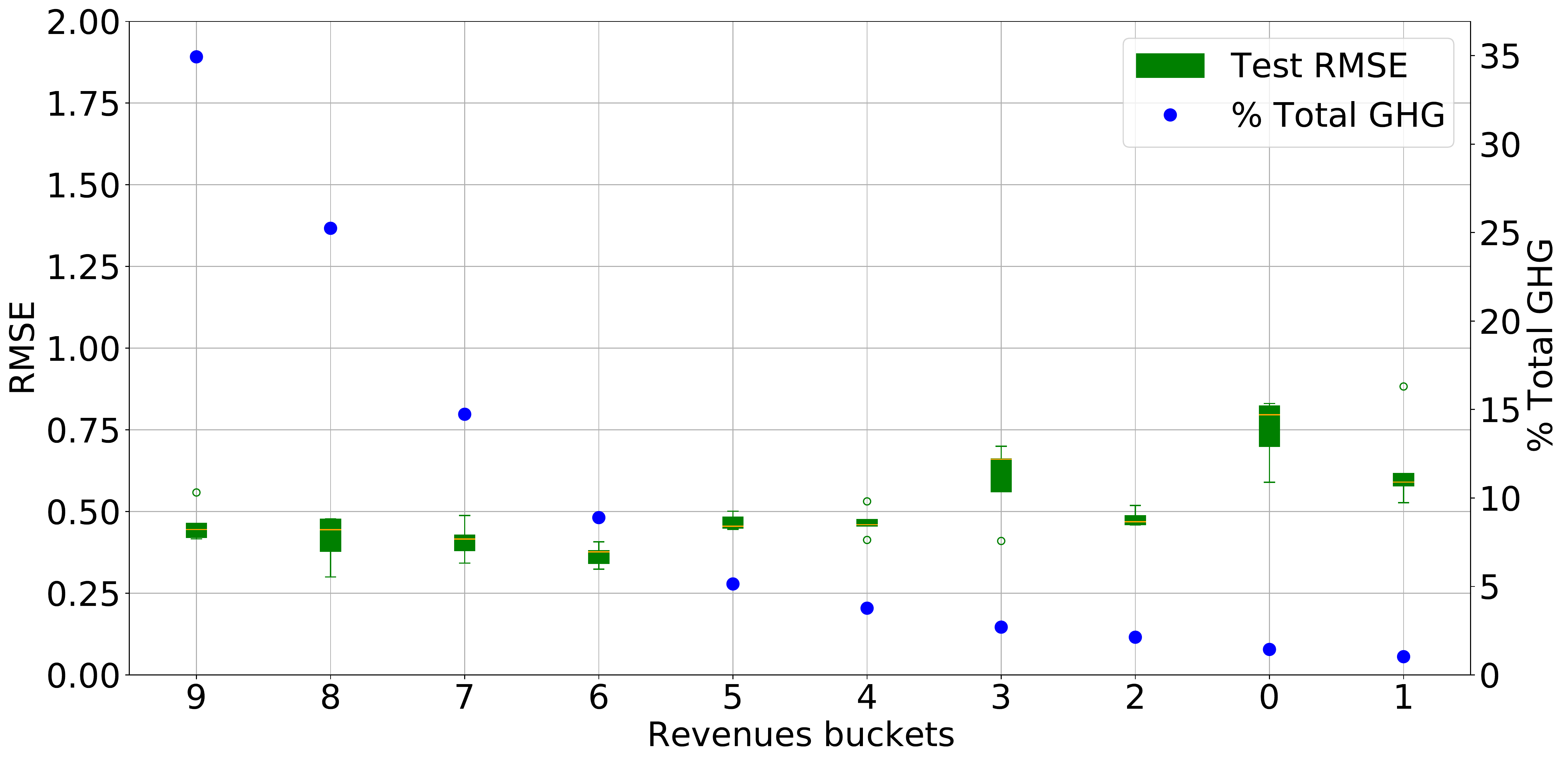}
		\caption{Boxplot of test RMSE on the five different test sets per deciles of revenues, ordered by revenues deciles emissions.}
		\label{fig:most_emissive_RMSE_CF2_2020_LGBM_BICS_Revenues Bucket_TPS_raw_v16_False}
	\end{subfigure}
	
	\caption{GHG emissions scope 2: distribution of performances of the model on five test sets according to different characteristics of companies.}
	\label{fig:ghg_emissions_scope2_summary_perf}
\end{figure}

Results are presented in Fig. \ref{fig:ghg_emissions_scope1_summary_perf} and \ref{fig:ghg_emissions_scope2_summary_perf} respectively for the scope 1 and scope 2.

Figures \ref{fig:most_emissive_lvl12_RMSE_CF1_2020_LGBM_BICS_BICS_TPS_raw_v16} and \ref{fig:most_emissive_lvl12_RMSE_CF2_2020_LGBM_BICS_BICS_TPS_raw_v16} show the RMSE distribution across the five test sets for BICS Sectors L1 (Level 1 of granularity) and L2 (Level 2 of granularity): the green box-plots correspond to the L2 sectors results and the pink ones in the background corresponds to the associated L1 sector. Results are ordered from the highest to the lowest emissivity of the BICS Sector L2, computed on the full set of reported data. This figures highlights that the model has rather stable performances across all sectors, with in particular good performances on the most emissive sectors. This plots also highlights the importance of the chosen sectorization methodology when evaluating a GHG model: sectors should regroup similar companies in terms of emissions. Knowing some sectors, like mining, gather sub-indutries with heterogenous GHG emissions schemes,  this could explain why our model currently have a bit more difficulties in estimating emissions for some sectors. For instance, in the mining sector, depending on the chosen technique, one ton of aluminium production can create around 10 times more emission than one ton of steel production.  The model performances for the most emissive BICS Sector L3 (Level 3 of granularity) is proposed in the Appendix \ref{app:perf_bics_l3}.

Figures \ref{fig:most_emissive_RMSE_CF1_2020_LGBM_BICS_IsoCtry_TPS_raw_v16_False} and \ref{fig:most_emissive_RMSE_CF2_2020_LGBM_BICS_IsoCtry_TPS_raw_v16_False} takes a similar approach by proposing the RMSE distribution across the five test sets per countries, for both scopes. Results are ordered by how emissive a country is in regard to the set of reported data. 

Finally, we show in Fig. \ref{fig:most_emissive_RMSE_CF1_2020_LGBM_BICS_Revenues Bucket_TPS_raw_v16_False} and \ref{fig:most_emissive_RMSE_CF2_2020_LGBM_BICS_Revenues Bucket_TPS_raw_v16_False} the RMSE performances across the five test sets per deciles of revenues. The 9\textsuperscript{th} decile of revenues corresponds to the one with the highest revenues, the 0\textsuperscript{th} is the one with the lowest. These graphs shows that in average, it is easier for the model to estimate the GHG emissions of companies with higher revenues. This may due to the fact that we have in our training sets more samples coming from big companies, as shown in Tab. \ref{tab:num_features_summary}, than ones coming from SMEs. Gathering more data from SMEs is a source of improvement for future versions of the model.

\section{Results: comparison of estimates with other providers}
\label{sec:results_providers}

To be more thorough, we now assess the quality of the estimates from our model, called the GHG-2022 model, in comparison to other data providers. These comparisons are done as of August 2022.

\paragraph{Retraining the model on the full dataset}

In section \ref{sec:results}, the models performances are evaluated using test sets. The samples in those test set could bring precious additional information to the model and should not be left aside in the final calibration of the model. Thus, to obtain the final model on which predictions will be made, we follow the procedure previously validated by the results in section \ref{sec:results} and train the models on the whole data, without test sets. Validation sets are still required to find the best models hyperparameters.

We consider the universe of 48429 companies extract from the Worldscope Refinitiv Database in 2020 to evaluate the prediction of our models and to compare them to other providers ones. 

\subsection{Comparison of coverage}

\begin{figure}
	\centering
	\includegraphics[width=1\textwidth]{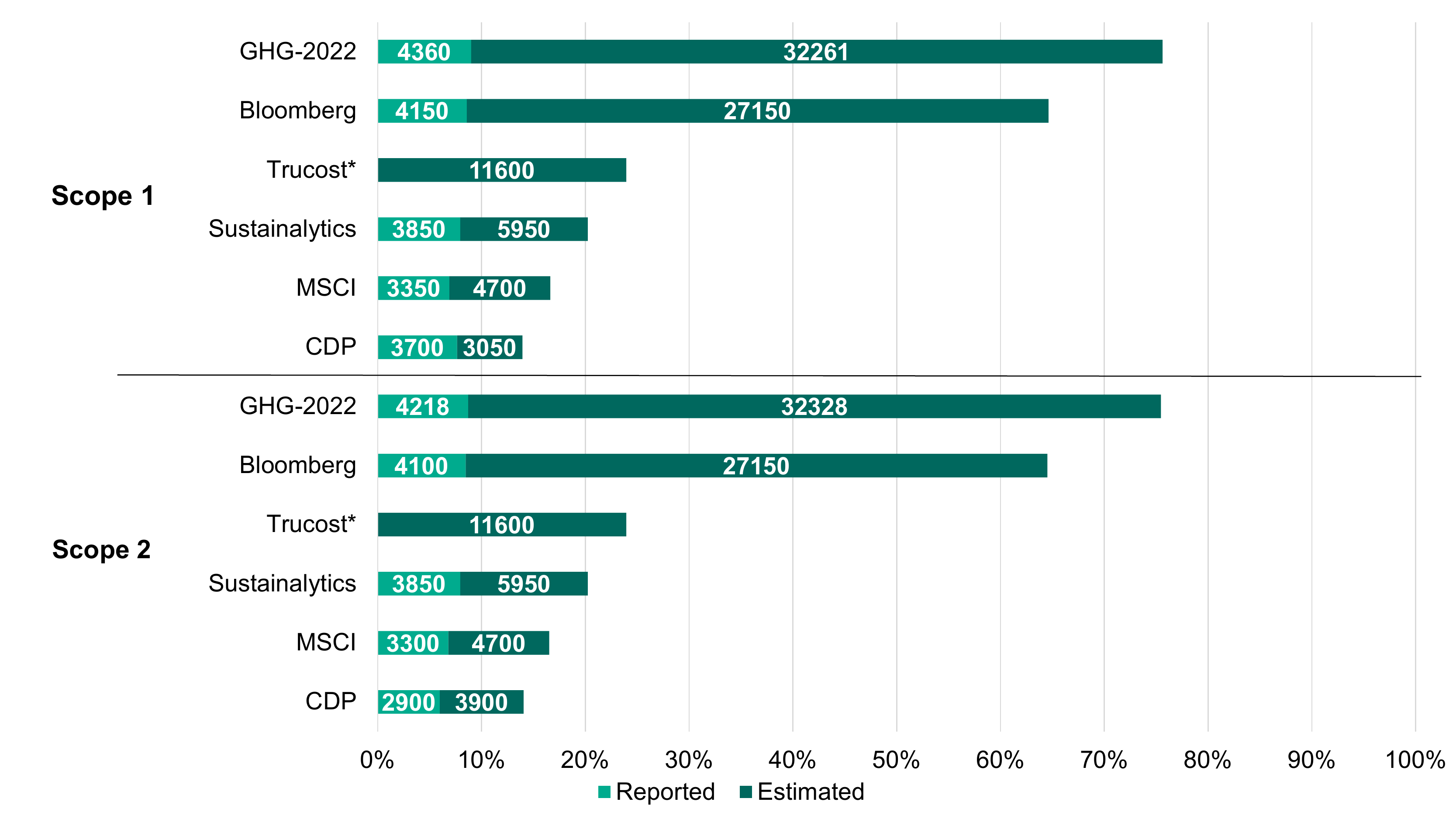}
	\caption{GHG emissions coverage, as of August 2022: number of reported data and estimates provided by each model. For the providers marked with an asterisk, the split between reported and estimated data was unclear, so all data points are marked as estimates.}
	\label{fig:Figure N Estimations}
\end{figure}

Figure \ref{fig:Figure N Estimations} displays for the scope 1 and scope 2 the number of reported GHG emissions and estimated GHG emissions each providers can provide for the year 2020. The test was conducted on the full universe of 48429 companies: for instance, our GHG-2022 model can provide for the scope 1 4360 reported data (sampled from CDP and Bloomberg and used for training as explained in section \ref{sec:raw_target_obtention}), and 32 261 estimates. For the remaining samples, the model was not able to provide an estimate mainly because of missing information for the company or because the considered values for categorical features were never seen during training: we do not want our model to extrapolate on categories unseen during calibration.

Coverages for Bloomberg, Trucost, Sustainalytics, MSCI and CDP are rounded as we may have slightly different results depending on the moment the datasets were obtained. Results provided in Fig. \ref{fig:Figure N Estimations} were obtained using the elements at our disposal in August 2022. 

Figure \ref{fig:Figure N Estimations} clearly demonstrates that using a machine learning model, fully automated and with a systematic methodology, allows to achieve an important coverage, greater than any other provider, while preserving good performances.

\subsection{Comparison of estimates accuracy}

To assess how good estimates are from one provider to another, we developed a methodology relying on the high year-over-year correlation of GHG estimates. The methodology is as follows: 
\begin{itemize}
	\item Using the same procedure, we train a model relying only on 2010 to 2018 data and a second one relying only on 2010 to 2019 data. These models, when used for predictions on 2018 and 2019 data respectively give 2018 and 2019 point-in-time estimates.
	\item We consider the reported values in 2020 for companies which started reporting in 2020 and thus have never reported in 2018 nor in 2019. This 2020 reported value is called the ground truth.
	\item  By comparing the 2019 estimates (or 2018 estimates if the 2019 is not available) from our GHG-2022 model and the ones from the other providers models to the 2020 ground truth, we can determine which provider is the closest to the ground truth and thus which provider seems to have the most accurate model. Comparison is done by computing a RMSE on the decimal logarithm of the estimation and ground truth.
\end{itemize}

Considering this methodology, we propose two ways of evaluating the providers:
\begin{itemize}
	\item First, we evaluate each of them separately. Tables \ref{tab:cf1_comp_all_points} and \ref{tab:cf2_comp_all_points} summarized these results for scope 1 and scope 2 respectively. The number of samples may greatly differ according to the coverage of the provider in estimates for companies which started reporting in 2020. Our model has the best, i.e. lowest, RMSE in comparison to the other considered providers.
	\item Second, we consider each provider against our GHG-2022 model. Results are available in tables \ref{tab:cf1_comp_common_points} and \ref{tab:cf2_comp_common_points} for scope 1 and scope 2 respectively. We rely this time on the same samples for GHG-2022 and the considered provider, increasing the comparability. In each case, we are systematically more accurate than the considered provider.
\end{itemize}

\paragraph{Point-in-time data} 
We train models using only 2018 and 2019 data respectively to avoid any leakage of the future in our 2018 and 2019 estimations. It may not be the case for the estimations of the other providers, which can bias the evaluation towards better performances of the other providers. The only provider for which we are sure estimates are done point-in-time is CDP. Even considering this, our model still have better performances than the considered providers. 

\begin{table}
\centering
	
\begin{subtable}{\textwidth}
	\centering
	\begin{tabular}{l|l|l}
		\textbf{Provider} & \textbf{RMSE} &  \textbf{Number of samples} \\
		\hline
		\hline
		GHG-2022 & 0.828 & 1079 \\
		\hline
		MSCI & 0.882 & 509 \\
		\hline
		Bloomberg & 0.948 & 1119 \\
		\hline
		Trucost & 1.033 & 980 \\
		\hline
		CDP & 1.222 & 546 \\
	\end{tabular}
	\caption{All companies are considered.}
	\label{tab:cf1_comp_all_points}
\end{subtable}

\begin{subtable}{\textwidth}
	\centering
	\begin{tabular}{l|l|l|l}
		\textbf{Provider} & \textbf{RMSE: provider} & \textbf{RMSE: GHG-2022} & \textbf{Number of samples} \\
		\hline
		\hline
		MSCI & 0.884 & 0.864 & 494 \\
		\hline
		Bloomberg & 0.956 & 0.828 & 1063 \\
		\hline
		Trucost & 1.039 & 0.812 & 952 \\
		\hline
		CDP & 1.228 & 0.849 & 530 \\
	\end{tabular}
	\caption{Only common companies between providers are considered.}
	\label{tab:cf1_comp_common_points}
\end{subtable}

\caption{Last scope 1 GHG estimates from providers (2019 – 2018) compare to 2020 ground truth, for companies which starts reporting in 2020.}
\label{tab:cf1_comp}
\end{table}

\begin{table}
\begin{subtable}{\textwidth}
	\centering
	\begin{tabular}{l|l|l}
		\textbf{Provider} & \textbf{RMSE} &  \textbf{Number of samples} \\
		\hline
		\hline
		GHG-2022 & 0.709 & 1042 \\
		\hline
		MSCI & 0.808 & 522 \\
		\hline
		Bloomberg & 0.809 & 1089 \\
		\hline
		Trucost & 0.822 & 955 \\
		\hline
		CDP & 0.970 & 577 \\
	\end{tabular}
	\caption{All companies are considered.}
	\label{tab:cf2_comp_all_points}
\end{subtable}

\begin{subtable}{\textwidth}
	\centering
	\begin{tabular}{l|l|l|l}
		\textbf{Provider} & \textbf{RMSE: provider} & \textbf{RMSE: GHG-2022} & \textbf{Number of samples} \\
		\hline
		\hline
		Bloomberg & 0.774 & 0.700 & 1029 \\
		\hline
		MSCI & 0.780 & 0.707 & 502 \\
		\hline
		Trucost & 0.803  & 0.645 & 925 \\
		\hline
		CDP & 0.950 & 0.661 & 561 \\
	\end{tabular}
	\caption{Only common companies between providers are considered.}
	\label{tab:cf2_comp_common_points}
\end{subtable}

\caption{Last scope 2 GHG estimates from providers (2019 – 2018) compare to 2020 ground truth, for companies which starts reporting in 2020.}
\label{tab:cf2_comp}
\end{table}

\paragraph{Breakdown of performances per sectors}

The methodology developed to compare our model to providers can be extended per sectors. In this section we only focus on the provider CDP as it is the only one for which we are sure estimates are done point-in-time, even if the coverage of CDP is relatively small compared to any other provider. 

\begin{figure}
	\centering
	\includegraphics[width=0.7\textwidth]{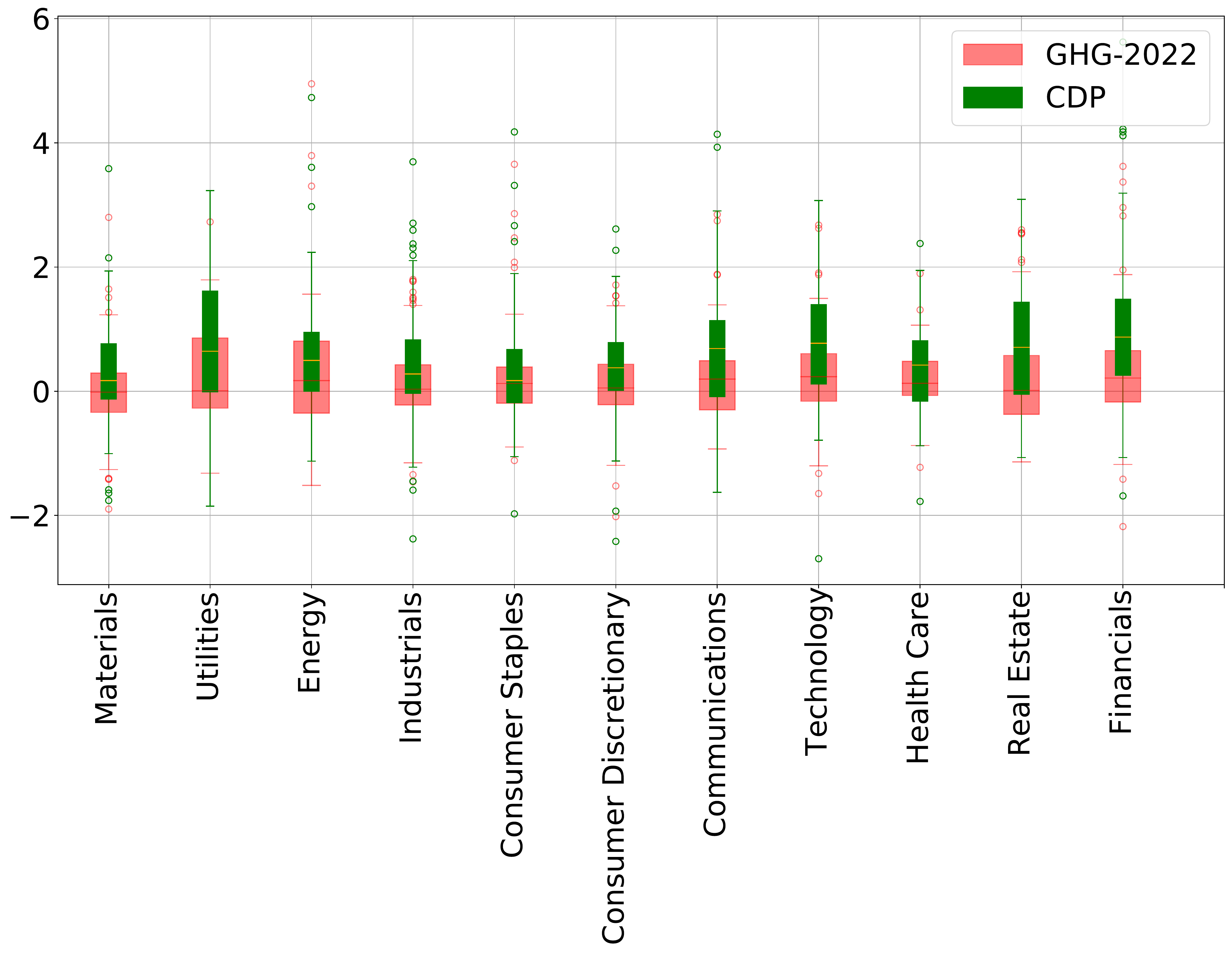}
	\caption{Differences of the emission from estimations from CDP and from GHG-2022 with ground truth for scopes 1 and 2.}
	\label{fig:CDP_comp_201819st_2020gt}
\end{figure}

For each sector of BICS level 1, we plot the distribution of the difference between the decimal logarithm of the ground truth and of the 2019 estimate (or 2018 if the 2019 one is not available) from the considered model. Results are displayed in Fig. \ref{fig:CDP_comp_201819st_2020gt}: in green, we observe the distribution of differences between CDP estimates and the ground truth, in pink, we observe the distribution of differences between our GHG-2022 estimates and the ground truth. Distributions from our model are more centered around 0, meaning better accuracy for our model than CDP. We can however notice than CDP estimates are more conservative than ours: when CDP estimates are not exact, they have a tendency to overestimate whereas our GHG-2022 model is rather balance between overestimation and underestimations. Both behavior and calibration from our model can have their strengths and weaknesses depending on the use case.

\section{Interpretability - Shapley values}

In the remainder of this work, we will provide tools to get insights on how the model works and why it estimates such values of GHG emissions. We provide a breakdown of the impact of the different training features on the estimated emissions from the models. A common critic of GBDT is that, despite their superior performances in tabular settings, they remain difficult to interpret. A tool, recently applied to the machine learning field and called Shapley Values, solves this issue.  

Shapley values, first introduced in the context of game theory \citep{shapley1953value}, provide a way in machine learning to characterize how each feature contributes to the formation of the final predictions. Shapley values and their uses in the context of machine learning are well described in \cite{molnar2020interpretable}.

The Shapley value of a feature can be obtained by averaging the difference of prediction between each combinations of features containing and not containing the said feature. For each sample in the dataset, each feature possesses its own Shapley value representing the contribution of this feature to the prediction for this particular sample. Shapley values have interesting properties, like the efficiency property. If we note $\phi_{j, i}$ the Shapley value of feature $j$ for a sample $x_i$ and $\hat{f}(x_i)$ the prediction for the sample $x_i$, Shapley values must add up to the difference between the prediction for the sample $x_i$ and the average of all predictions $\mathbb{E}_X(\hat{f}(X))$ and then follow the following formula:

\begin{align}
	\sum\nolimits_{j=1}^p\phi_j=\hat{f}(x)-\mathbb{E}_X(\hat{f}(X))
\end{align}

The dummy property states that the Shapley value of a feature which does not change the prediction, whatever combinations of features it is added to, should be 0. 

Shapley values calculation is quite time and memory intensive. \cite{lundberg2017unified} and later \cite{lundberg2018consistent} proposed an implementation of a fast algorithm called TreeSHAP, which allows to approximate Shapley values for trees models like the LightGBM, which we use in the following and refer to as SHAP values.

\subsection{SHAP feature importance}

\begin{figure}
	\centering
	\begin{subfigure}{0.49\textwidth}
		\centering
		\includegraphics[width=1\textwidth]{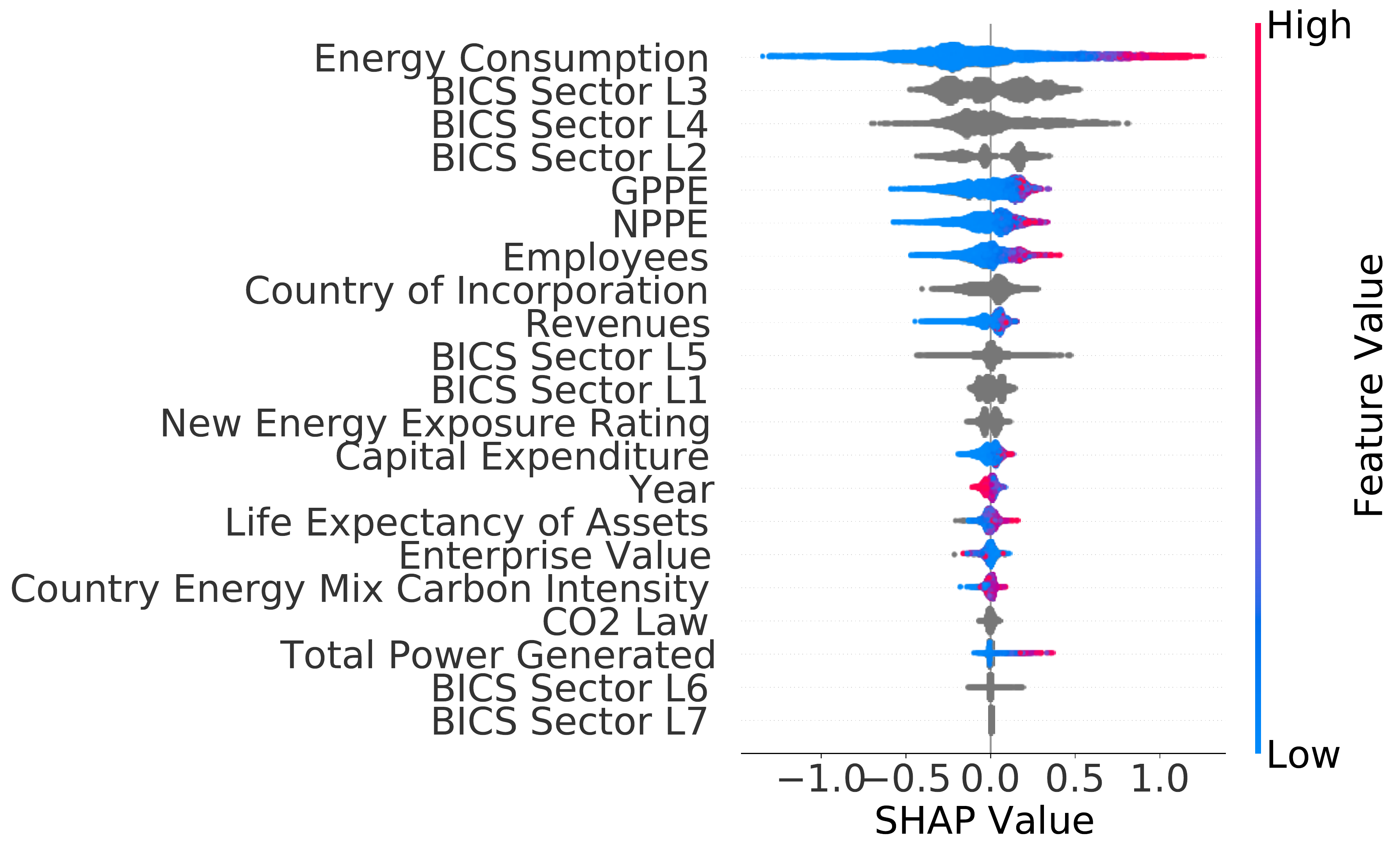}
		\caption{Scope 1}
		\label{fig:shap_CF1_beeswarm}
	\end{subfigure}
	\hfill
	\begin{subfigure}{0.49\textwidth}
		\centering
		\includegraphics[width=1\textwidth]{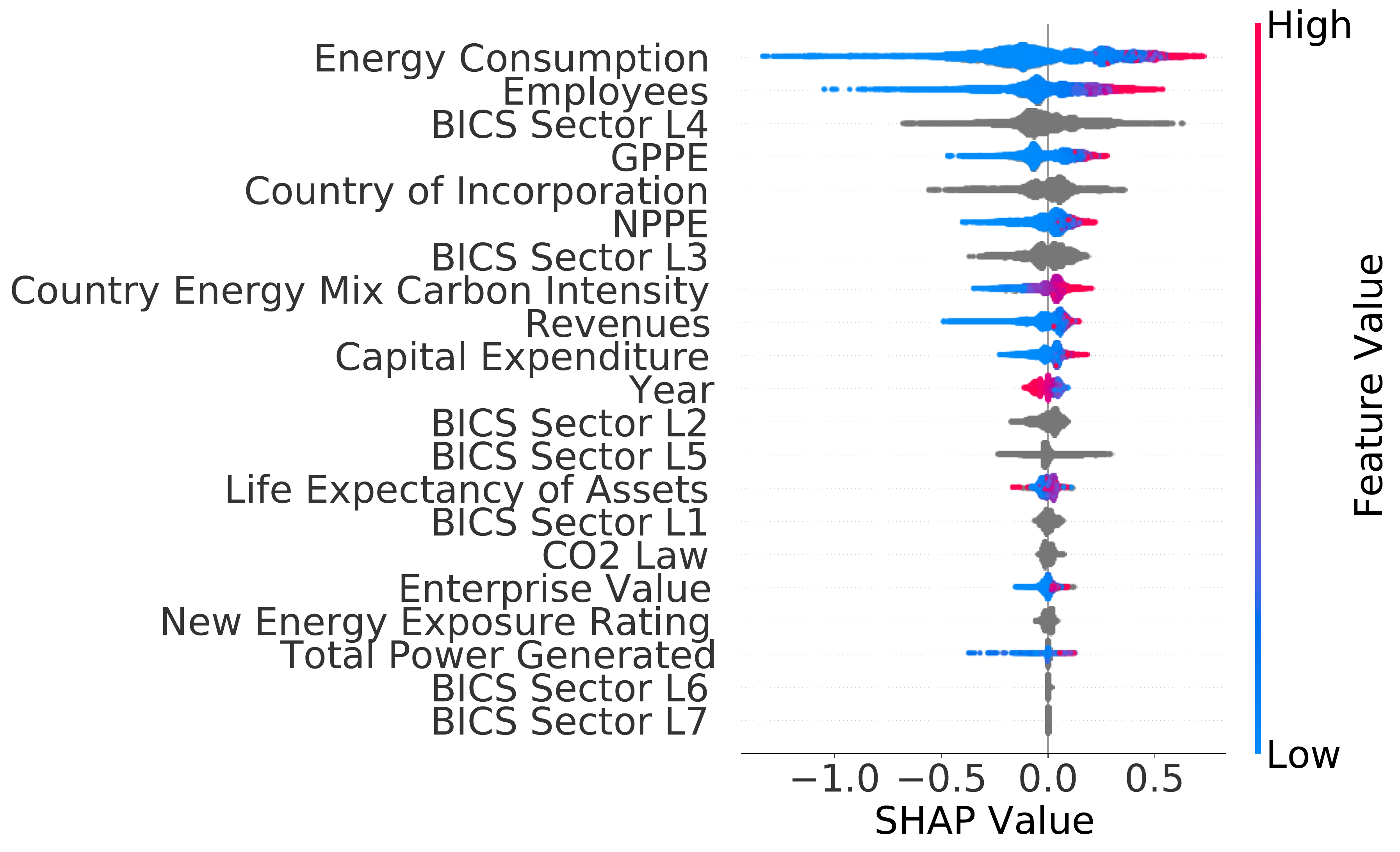}
		\caption{Scope 2}
		\label{fig:shap_CF2_beeswarm}
	\end{subfigure}
	
	\caption{SHAP values: impact of each feature on the predicted GHG emission, order by importance.}
	\label{fig:shap_CF_beeswarm}
\end{figure}

We provide in Fig. \ref{fig:shap_CF_beeswarm} the breakdown of SHAP values per features for the scope 1 and scope 2 GHG emissions, ordered by importance. For each feature, this graph shows the distribution of SHAP values across each samples in our training set. These graphs are key elements in our model as they make it interpretable: they can be computed for any set of features, allowing to understand why the model make a specific decision and output this predicted estimate. 

The Energy Consumption feature is the most important one used by the model for both scope 1 and scope 2, which seems logic. Industry classification features are also very important for both scopes. As we could have expected from the definition of scope 2, the features Employees, Country of Incorporation and Country Energy Mix Carbon Intensity are more important for the estimation of scope 2 than the estimation of scope 1. 

Knowing these SHAP values not only allows to better understand an estimate the model output but also to evaluate the reliability of the estimate based on the presence or the absence of a feature: if the Energy Consumption feature is not given for a sample, for certain sectors, it would lead to a less reliable estimate.

\subsection{Relationship between features values and GHG estimates}
\label{sec:relfeatshap}

\begin{figure}
	\centering
	\begin{subfigure}{0.48\textwidth}
		\centering
		\includegraphics[width=1\textwidth]{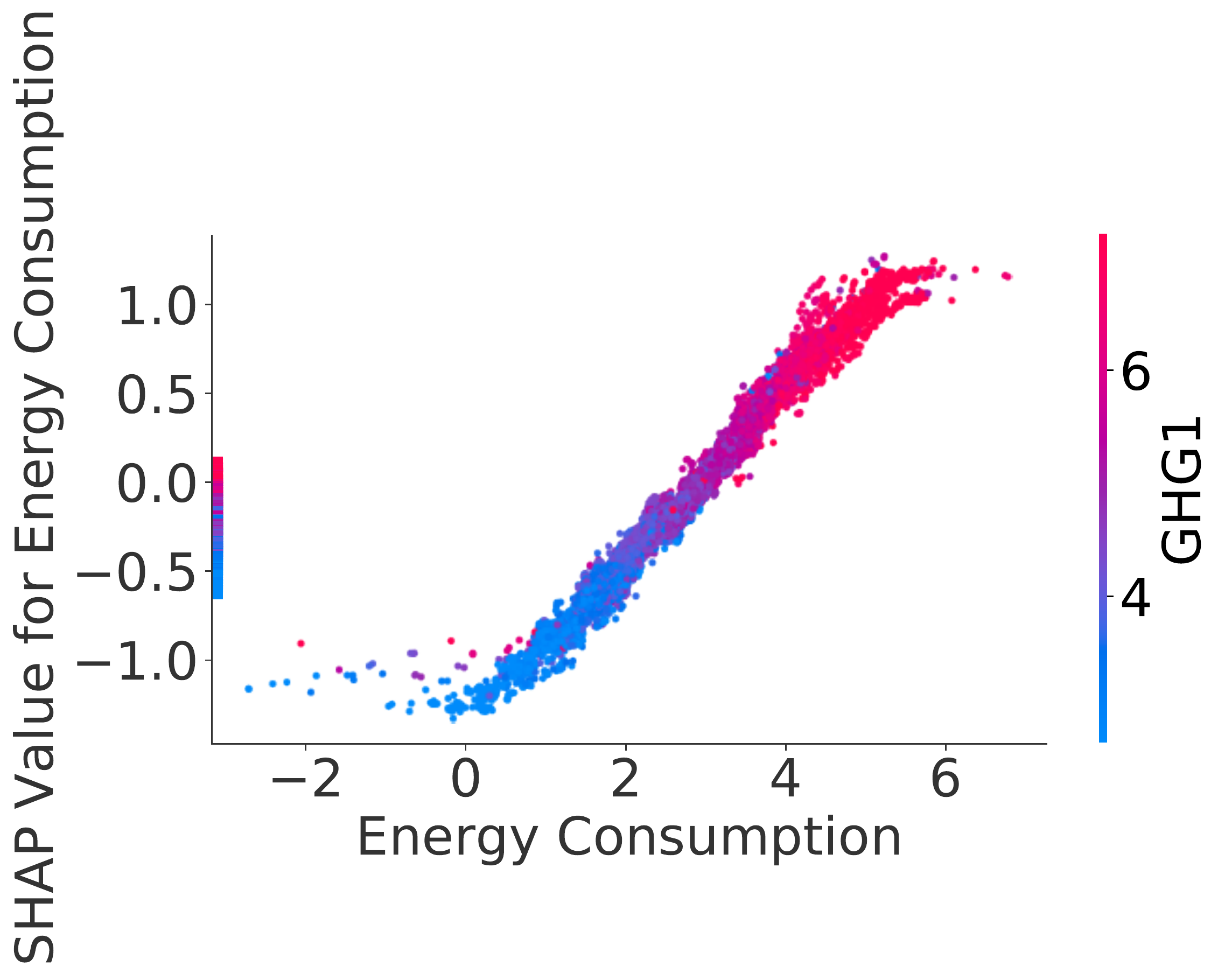}
		\caption{Scope 1}
		\label{fig:shap_CF1_Energy Consumption}
	\end{subfigure}
	\hfill
	\begin{subfigure}{0.48\textwidth}
		\centering
		\includegraphics[width=1\textwidth]{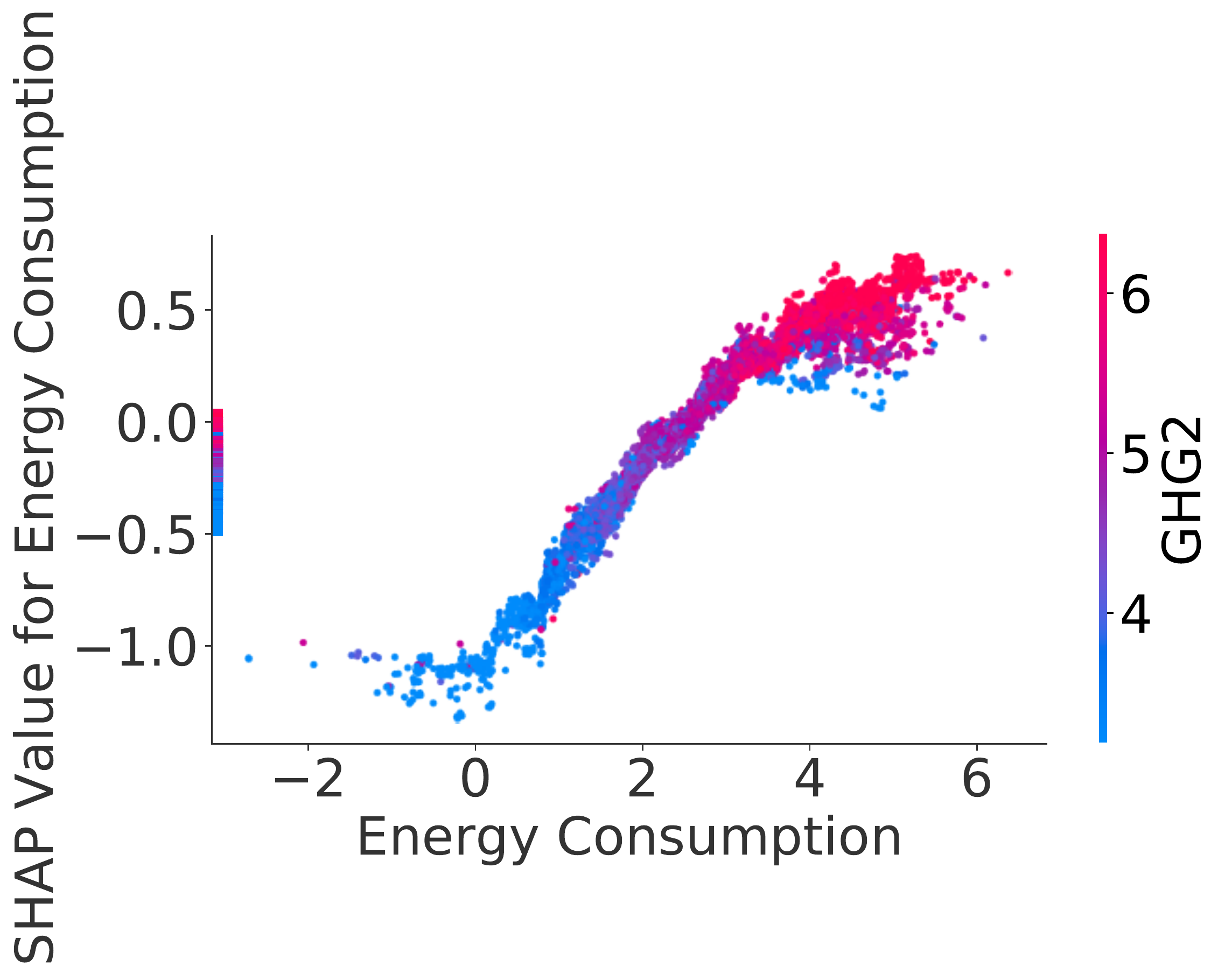}
		\caption{Scope 2}
		\label{fig:shap_CF2_Energy Consumption}
	\end{subfigure}
	\caption{Relationship between SHAP values of the Energy Consumption feature and the decimal logarithm of the Energy Consumption feature value.}
	\label{fig:shap_CF_Energy Consumption}
\end{figure}

\paragraph{Numerical features}
SHAP values can be computed for each feature on each sample. This tool allows to better understand the relationship captured by the model between a feature and the estimated GHG emission. Indeed, for numerical features, we can plot the SHAP values for a specific feature against this feature value in the dataset. For instance, Fig. \ref{fig:shap_CF_Energy Consumption} shows the relation between SHAP values of the Energy Consumption feature and the decimal logarithm of the Energy Consumption feature value. Apart from the points which are on the Y-axis and which represents missing values for the Energy Consumption feature, there is for both scopes a near-linear increasing relationship between the SHAP values of the Energy Consumption feature and the decimal logarithm of this feature value. Appendices  \ref{app:shap_numerical_employees_revenues} provides some others examples of  SHAP values plots for numerical features, allowing for a better interpretation of what the model is learning.

\begin{figure}
	\centering
	\begin{subfigure}{0.8\textwidth}
		\centering
		\includegraphics[width=1\textwidth]{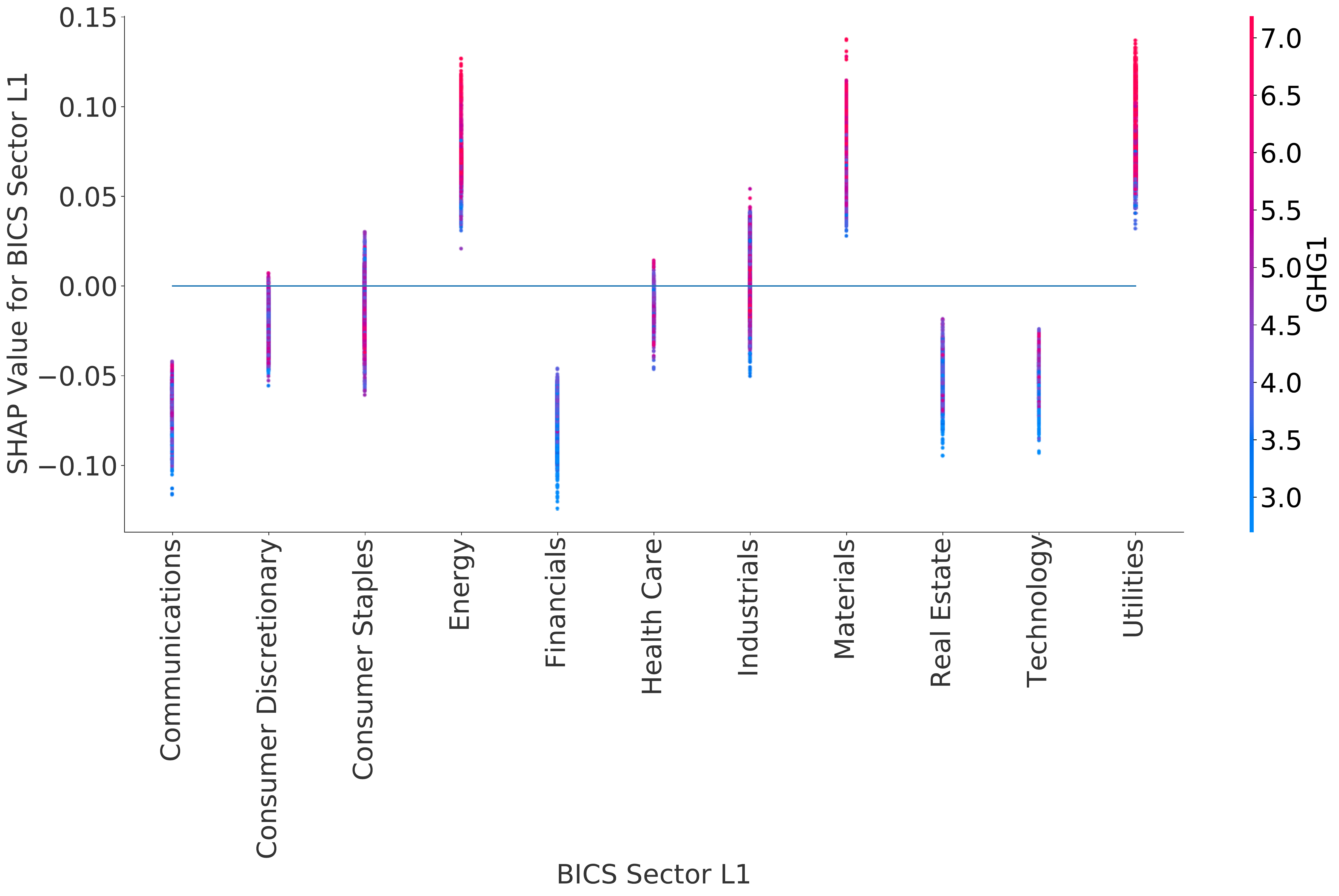}
		\caption{Scope 1}
		\label{fig:shap_CF1_BICS Sector L1}
	\end{subfigure}
	
	\begin{subfigure}{0.8\textwidth}
		\centering
		\includegraphics[width=1\textwidth]{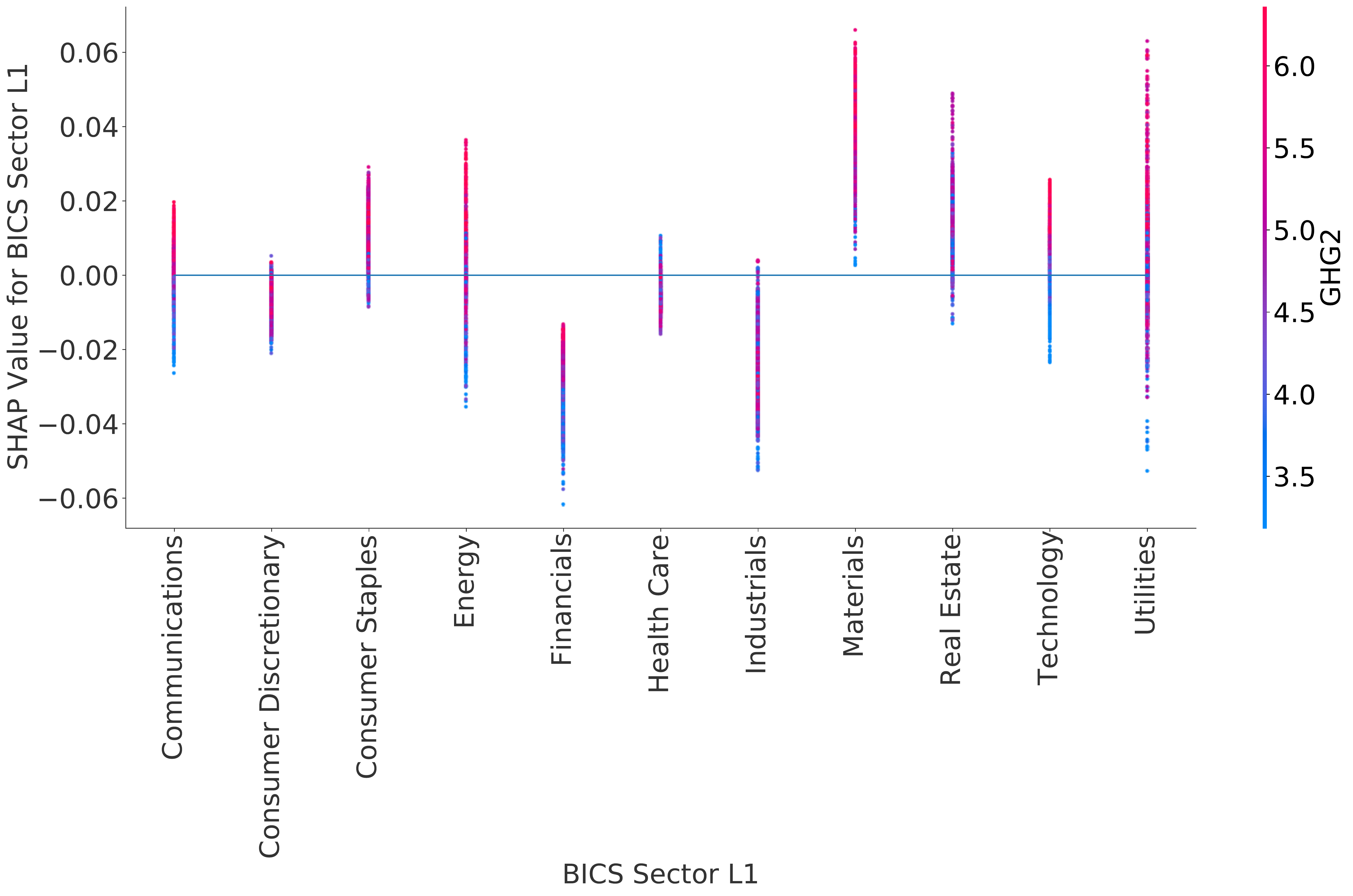}
		\caption{Scope 2}
		\label{fig:shap_CF2_BICS Sector L1}
	\end{subfigure}
	
	\caption{SHAP values: impact of belonging to a particular level 1 BICS sector on the predicted GHG emission.}
	\label{fig:shap_CF_BICS Sector L1}
\end{figure}

\paragraph{Categorical features}
SHAP values can also be used on categorical features to study their distribution, for each value a categorical feature can take. For instance,
Fig. \ref{fig:shap_CF_BICS Sector L1} shows the distribution of SHAP values for the BICS Sector L1 feature, for each of the BICS Sector L1 sectors. Thanks to this plot, we can understand in what sectors companies are more likely to have higher GHG emissions. For instance, for scope 1, SHAP values for all companies in the Energy and Materials sectors show an increase to the estimated emission (positive SHAP values). On the contrary, samples in the Financial sector have negative SHAP values, showing a decrease to the estimated emission.

This plot can be done for all categorical features, allowing to understand the distribution of SHAP values according to each category, and then having a better interpretation of the model. In Appendix \ref{app:shap_categorical_year}, we provide an additional SHAP values plot for categorical features for further interpretation elements.

Plots in Fig. \ref{fig:shap_CF_BICS Sector L1} highlight some clusters of SHAP values inside the distribution of BICS Sector L1 SHAP values per BICS Sector L1. These clusters show differences in the distribution of the initial data. Working on these clusters and removing the ones with too few samples could be a solution to improve the model by removing outliers and preventing overfitting. For instance, working with the distribution of SHAP values for the Utilities sector in scope 1, we see that there is a cluster of SHAP values below 0.04 with few samples. These corresponds to the years 2012 to 2014 of a specific company for which the reported Energy Consumption is around 19 000 GWh whereas the reported values for the same company from 2015 to 2020 are between 30 and 65 GWh. The removal of this cluster with very few samples allows us to improve the quality of the training data by removing outliers. Similar studies on other sectors leads to the same results: for the Materials sector, it can lead to the removal of the only years a company did not report its Energy Consumption for instance. 

\paragraph{Data Polishing}
This methodology should however be automatized and applied systematically. We did a first implementation using the SHAP distribution for each BICS Sector L4 (Level 4 of granularity). For each L4 sector, we applied a hierarchical clustering algorithm, separating clusters if their distance is above 0.04 in the SHAP values space, and remove clusters of data with an insufficient number of sample, i.e. less than 10. All these parameters were found by trials-and-errors. For both scope 1 and scope 2, it leads to the removal of about respectively 11.5\% and 5\% of the training data, enabling an improvement of the global performance of the model. Results are presented in Tab. \ref{tab:data_pol_global_metrics_results}, on average on the 5 different test sets: we observe for both scopes an average RMSE decrease between 11\% and 13\%. Let's however note that it may come at the price of an increase of variability between results in the different test set, especially for scope 1, as shown by the standard deviation of the RMSE metric in Tab. \ref{tab:scope1_data_pol_global_metrics_results}. As future work, studying more the impact of this methodology on both global and granular performances may lead to a more accurate and robust model. 

\begin{table}
\centering	

\begin{subtable}{\textwidth}
	\centering
	\begin{tabular}{l l||c|c|c|c}
		& & \multicolumn{2}{c|}{\textbf{Without data polishing}} & \multicolumn{2}{c}{\textbf{With data polishing}} \\
		\hline
		\textbf{Range} & \textbf{Metric} & \textbf{Mean} & \textbf{Standard Deviation} & \textbf{Mean} & \textbf{Standard Deviation}\\
		\hline
		\hline
		$[0, 1]$ & $R^2$ & 0.832 & 0.007 & 0.859 & 0.009 \\
		\hline
		\hline
		$[0, +\inf[$ & RMSE & 0.578 & 0.007  & 0.501 & 0.020 \\
		\hline
		\hline
		$[0, +\inf[$ & MAE & 0.401 & 0.006 & 0.347 & 0.013 \\
	\end{tabular}
	\caption{Scope 1}
	\label{tab:scope1_data_pol_global_metrics_results}
\end{subtable}

\begin{subtable}{\textwidth}
	\centering
	\begin{tabular}{l l||c|c|c|c}
		& & \multicolumn{2}{c|}{\textbf{Without data polishing}} & \multicolumn{2}{c}{\textbf{With data polishing}} \\
		\hline
		\textbf{Range} & \textbf{Metric} & \textbf{Mean} & \textbf{Standard Deviation} & \textbf{Mean} & \textbf{Standard Deviation}\\
		\hline
		\hline
		$[0, 1]$ & $R^2$ & 0.746 & 0.017 & 0.778 & 0.017 \\
		\hline
		\hline
		$[0, +\inf[$ & RMSE & 0.521 & 0.031  & 0.464 & 0.025 \\
		\hline
		\hline
		$[0, +\inf[$ & MAE & 0.341 & 0.010 & 0.312 & 0.011 \\
	\end{tabular}
	\caption{Scope 2}
	\label{tab:scope2_data_pol_global_metrics_results}
\end{subtable}

\caption{Results of the model on five different test sets, without and with the data polishing methodology applied: mean and standard deviation of the $R^2$, RMSE and MAE metrics. The three metrics, computed on the decimal logarithm of the emissions, are given for comparability purposes across the literature and should not be compared to each other.}
\label{tab:data_pol_global_metrics_results}
\end{table}

\section{Conclusion}

To mitigate the fact that GHG emissions reporting and auditing are not yet compulsory for all companies and that methodologies of measurement and estimations are not unified, we proposed a machine learning model to predict non-reported company emissions for scopes 1 and 2. Our model showed good out-of-sample performances when assessing it globally as well as good and balanced out-of-sample performances when assessing it per sector, per country and per buckets of revenues. Comparing our results to those of other providers, we found, as of August 2022, our estimates to be available for a larger number of companies and more accurate.

In addition to the large coverage and the accuracy, this model is also flexible, allowing for easy evolution in the input data as regulations evolve. We built a transparent and explainable model: the methodology is described in this study extensively, and the implemented tools based on Shapley values allow to understand the role played by each feature in the construction of the final output. We focused on some interpretability elements we found particularly interesting. Many more interactions could have been studies: interaction between sectors, revenues and the estimated GHG emission, redoing the study done in this paper for each sector separately... These would give even more information on how the model is working and allow to understand the specificities of GHG emissions per sectors. Studying all these SHAP values interactions is beyond the scope of this study and could be the object of an entire future publication. 

Future work to improve our model will first focus on gathering and including more training data from SMEs, so that the coverage of our model can be further improve. As we stressed in our analysis, the used industry classification is critical. We will gather data about all the activities a company is reporting being active in, and work on including this new industry classification in our model: it will help in improving accuracy in industries where companies operating in very different sectors in terms of GHG emissions have been grouped. The data polishing method that we introduced in the last section of this study will also be developed with the goal of obtaining a more robust model.

\pagebreak
\clearpage

\section*{Funding}

This work originates from a partnership between CentraleSupélec, Université Paris-Saclay and BNP Paribas.
This research received no external funding.

\section*{Author Contributions}

Conceptualization, J.A. and T.H.; Methodology, J.A., T.H. and L.C.; Software, J.A.; Validation, J.A. and T.H.; Formal Analysis, J.A. and T.H.; Investigation,  J.A. and T.H.; Resources, T.H. and L.C.; Data Curation, J.A. and T.H.; Writing – Original Draft Preparation, J.A. and T.H.; Writing – Review \& Editing: J.A., T.H., L.C. and F.S.; Visualization, J.A.; Supervision, L.C. and F.S.; Project Administration, J.A., T.H., L.C. and F.S.

\section*{Data Availability Statement}

Restrictions apply to the availability of some data. 
Data obtained from Refinitiv, Bloomberg, MSCI, Trucost and CDP are available from the authors with the permission of respectively Refinitiv, Bloomberg, MSCI, Trucost and CDP.

Data from WorkldBank and the International Energy Agency are publicly accessible and can be found respectively here \url{https://carbonpricingdashboard.worldbank.org} and here \url{https://api.iea.org/stats/indicators/CO2Intensity}.

\section*{Conflicts of Interest}

The authors declare no conflict of interest.

\pagebreak
\clearpage

\section*{Abbreviations}

\begin{tabular}{@{}ll}
BICS & Bloomberg Industry Classification Standard\\
BICS Sector L1 & BICS Sector Level 1, first level of granularity\\
BICS Sector L2 & BICS Sector Level 2, second level of granularity\\
BICS Sector L3 & BICS Sector Level 3, third level of granularity\\
BICS Sector L4 & BICS Sector Level 4, fourth level of granularity\\
CCS & Carbon Capture and Storage\\
CDP & Carbon Disclosure Project\\
COP & Conferences Of the Parties\\
CSR & Corporate Social Responsibility\\
CSRD & Corporate Sustainability Reporting Directive\\
EIO & Environmental Input Output\\
ESG & Environment, Social and Governance\\
ETS & Emission Trading System\\
EU & European Union\\
GBDT & Gradient Boosted Decision Trees\\
GGLR & Gamma Generalized Linear Regression\\
GHG & Greenhouse Gases\\
GPPE & Gross Property Plant \& Equipment\\
GWP & Global Warming Potential\\
IEA & International Energy Agency\\
MAE & Mean Absolute Error\\
MMD & Mean Maximum Discrepancy\\
MSE & Mean-Squared Error\\
NPPE & Net Property Plant \& Equipment\\
NZBA & Net Zero Banking Alliances\\
OLS & Ordinary Least Squares\\
PA & Process Analysis\\
PACTA & Paris Agreement Capital Transition Assessment\\
PCAF & Partnership for Carbon Accounting Financials\\
RMSE & Root Mean-Squared Error\\
SEC & Securities and Exchange Commission\\
SME & Small and Medium-sized Enterprise\\
WBCSD & World Business Council for Sustainable Development\\
WRI & World Resources Institute\\
\end{tabular}

\pagebreak
\clearpage

\bibliographystyle{plainnat}
\bibliography{reference}

\pagebreak
\clearpage

\appendix

\section{Model performances: BICS Sectors Level 3}
\label{app:perf_bics_l3}

\begin{figure}
	\centering
	\begin{subfigure}{0.65\textwidth}
		\centering
		\includegraphics[width=1\textwidth]{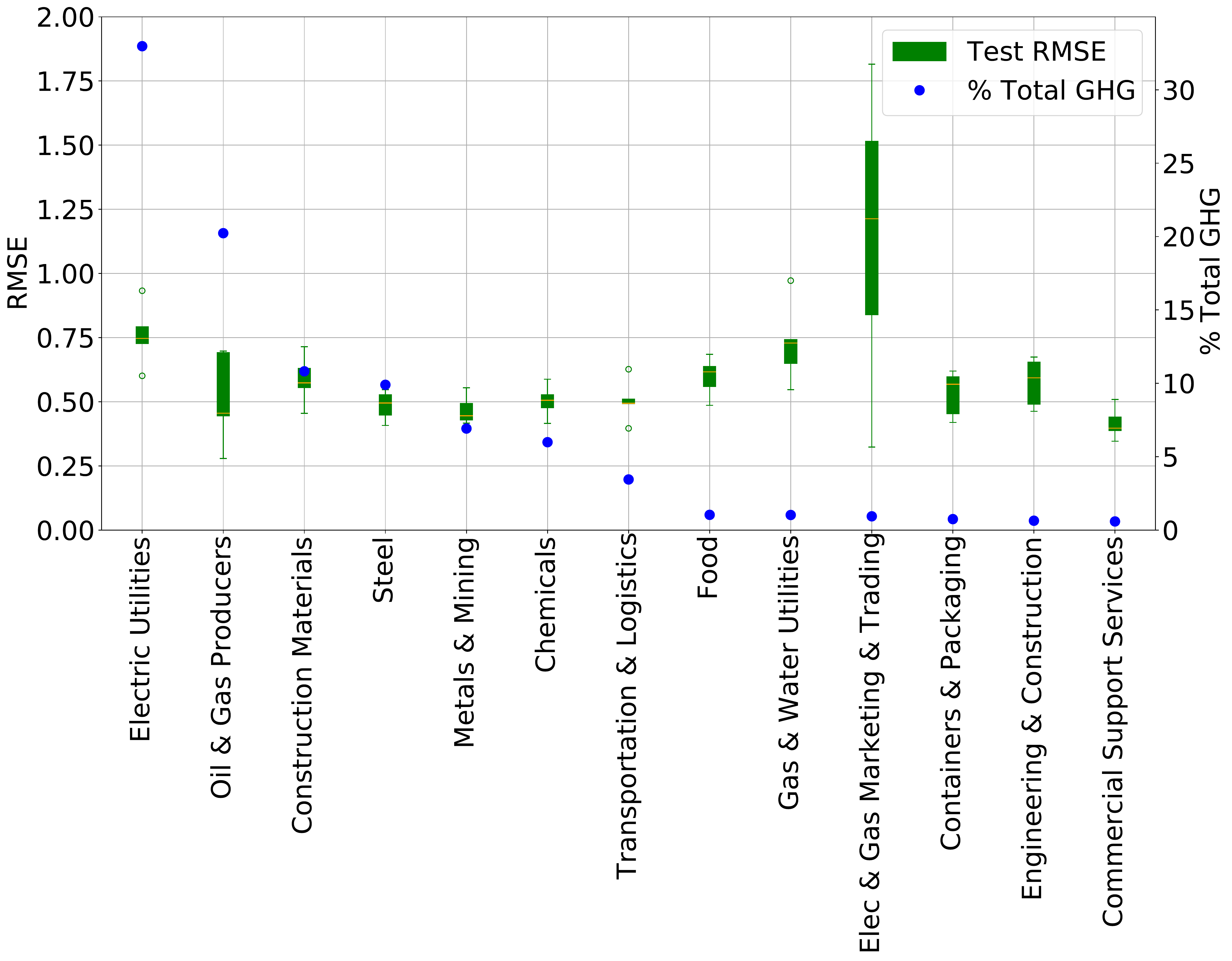}
		\caption{Scope 1}
		\label{fig:most_emissive_RMSE_CF1_2020_LGBM_BICS_BICS Sector L3_TPS_raw_v16_False}
	\end{subfigure}
	
	\begin{subfigure}{0.65\textwidth}
		\centering
		\includegraphics[width=1\textwidth]{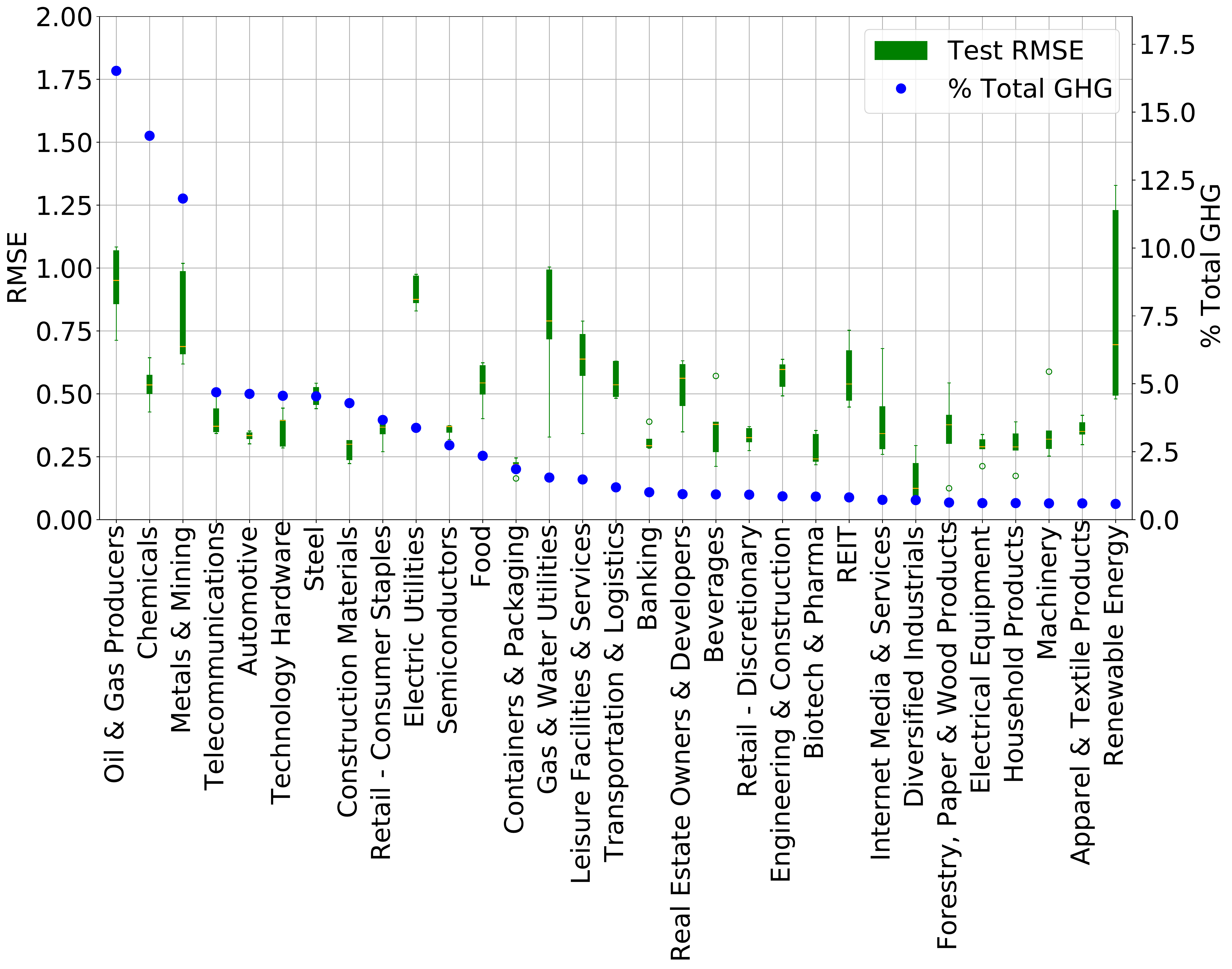}
		\caption{Scope 2}
		\label{fig:most_emissive_RMSE_CF2_2020_LGBM_BICS_BICS Sector L3_TPS_raw_v16_False}
	\end{subfigure}
		
	\caption{GHG emissions: distribution of performances of the model on five test sets according to BICS sectors level 3.}
	\label{fig:perf_bics_l3}
\end{figure}

In addition to the plots displayed in Fig. \ref{fig:most_emissive_RMSE_CF1_2020_LGBM_BICS_BICS Sector L3_TPS_raw_v16_False} and \ref{fig:most_emissive_RMSE_CF2_2020_LGBM_BICS_BICS Sector L3_TPS_raw_v16_False}, we provide for transparency purposes the breakdown of the out-of-sample performances of our model across the five test sets for the different BICS Sector L3, ranked from high to low emissivity, for sectors accounting for at east 1\% of the total GHG emissions of the reporting companies. 

\section{Relationship between features value and GHG estimates}

In addition to the plots and interpretation elements provided in section \ref{sec:relfeatshap}, we show some additional results for a better understanding of the learned relationships in the model.

\subsection{Numerical Data}
\label{app:shap_numerical_employees_revenues}

\begin{figure}
	\centering
	\begin{subfigure}{0.49\textwidth}
		\centering
		\includegraphics[width=1\textwidth]{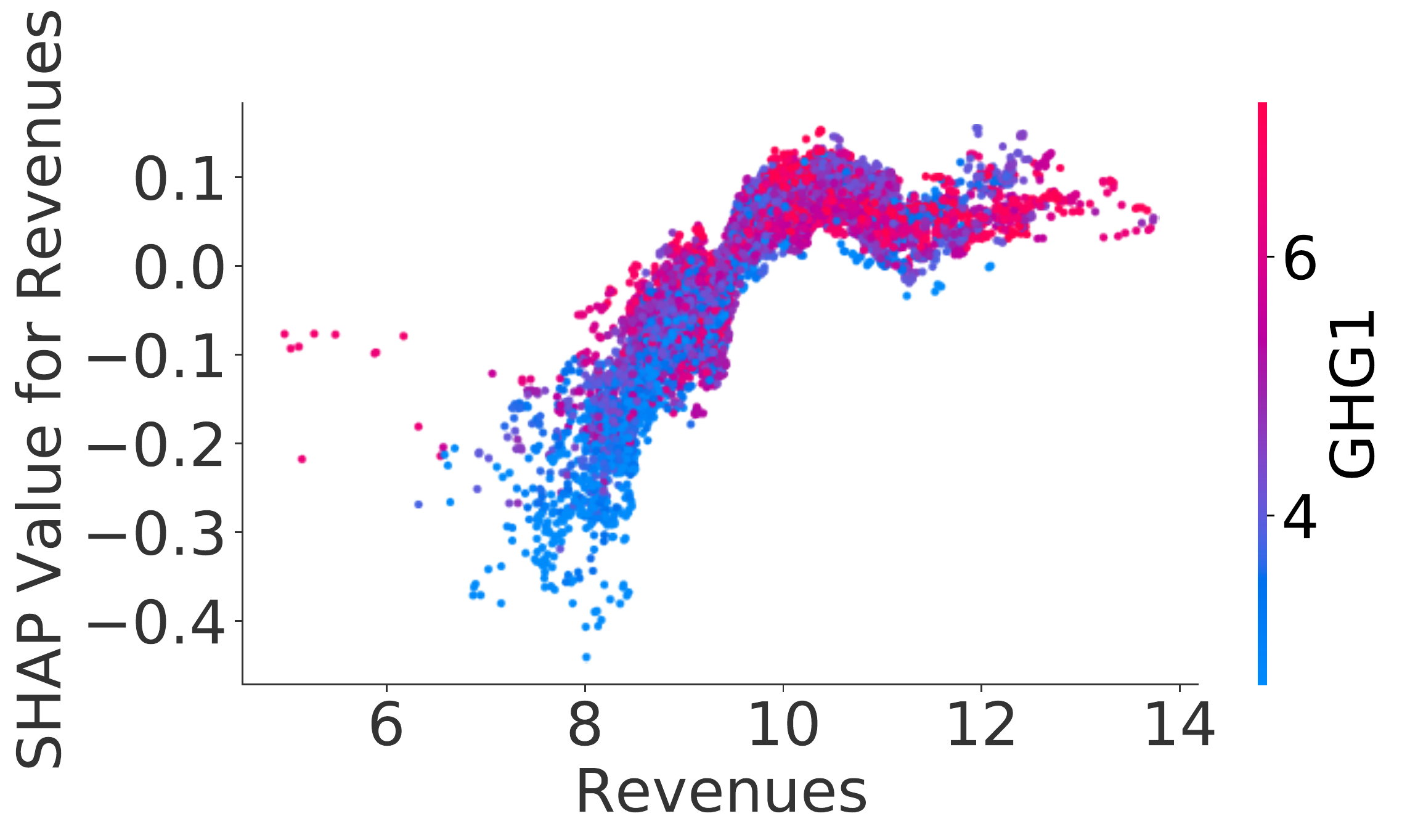}
		\caption{Scope 1}
		\label{fig:shap_CF1_Revenues}
	\end{subfigure}
	\hfill
	\begin{subfigure}{0.49\textwidth}
		\centering
		\includegraphics[width=1\textwidth]{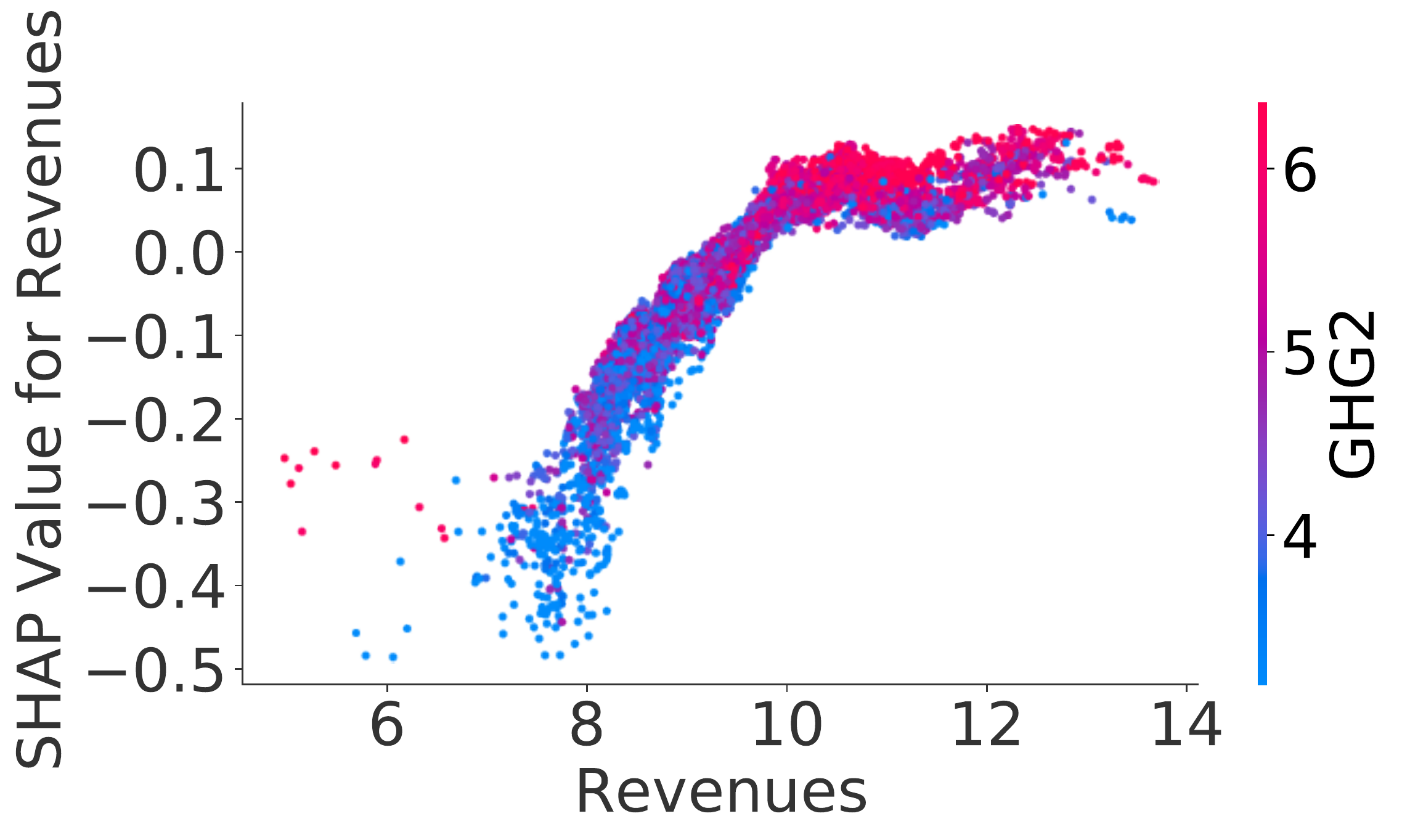}
		\caption{Scope 2}
		\label{fig:shap_CF2_Revenues}
	\end{subfigure}
	\caption{Relationship between SHAP values of the Revenues feature and the decimal logarithm of the Revenues feature value.}
	\label{fig:shap_CF_Revenues}
\end{figure}

Figure \ref{fig:shap_CF_Revenues} shows a near-linear relationship between the SHAP values of the Revenues feature and the decimal logarithm of the Revenues feature values, until a sort of cap: beyond a certain revenue level, the SHAP values are almost constant. 

\begin{figure}
	\centering
	\begin{subfigure}{0.49\textwidth}
		\centering
		\includegraphics[width=1\textwidth]{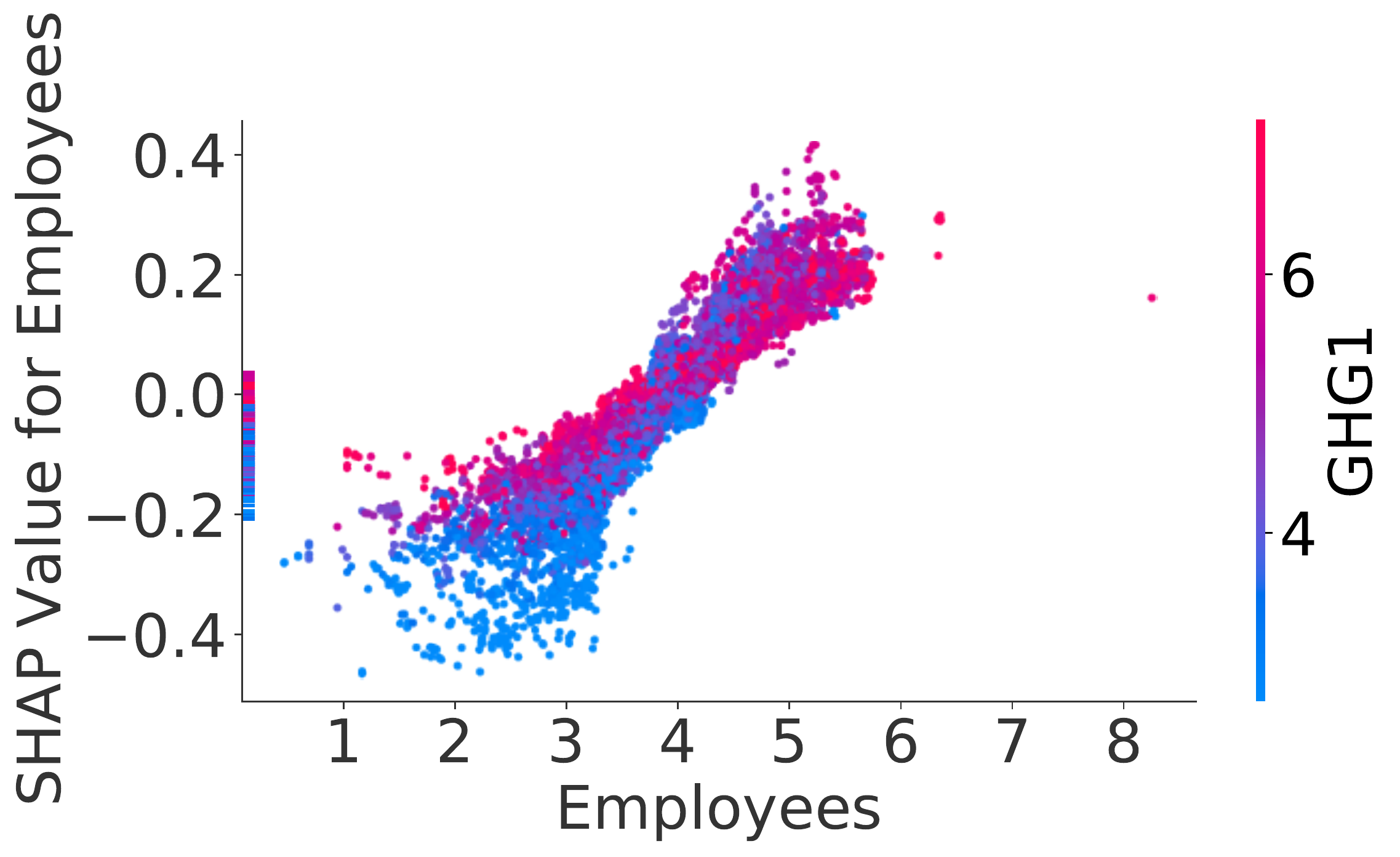}
		\caption{Scope 1}
		\label{fig:shap_CF1_Employees}
	\end{subfigure}
	\hfill
	\begin{subfigure}{0.49\textwidth}
		\centering
		\includegraphics[width=1\textwidth]{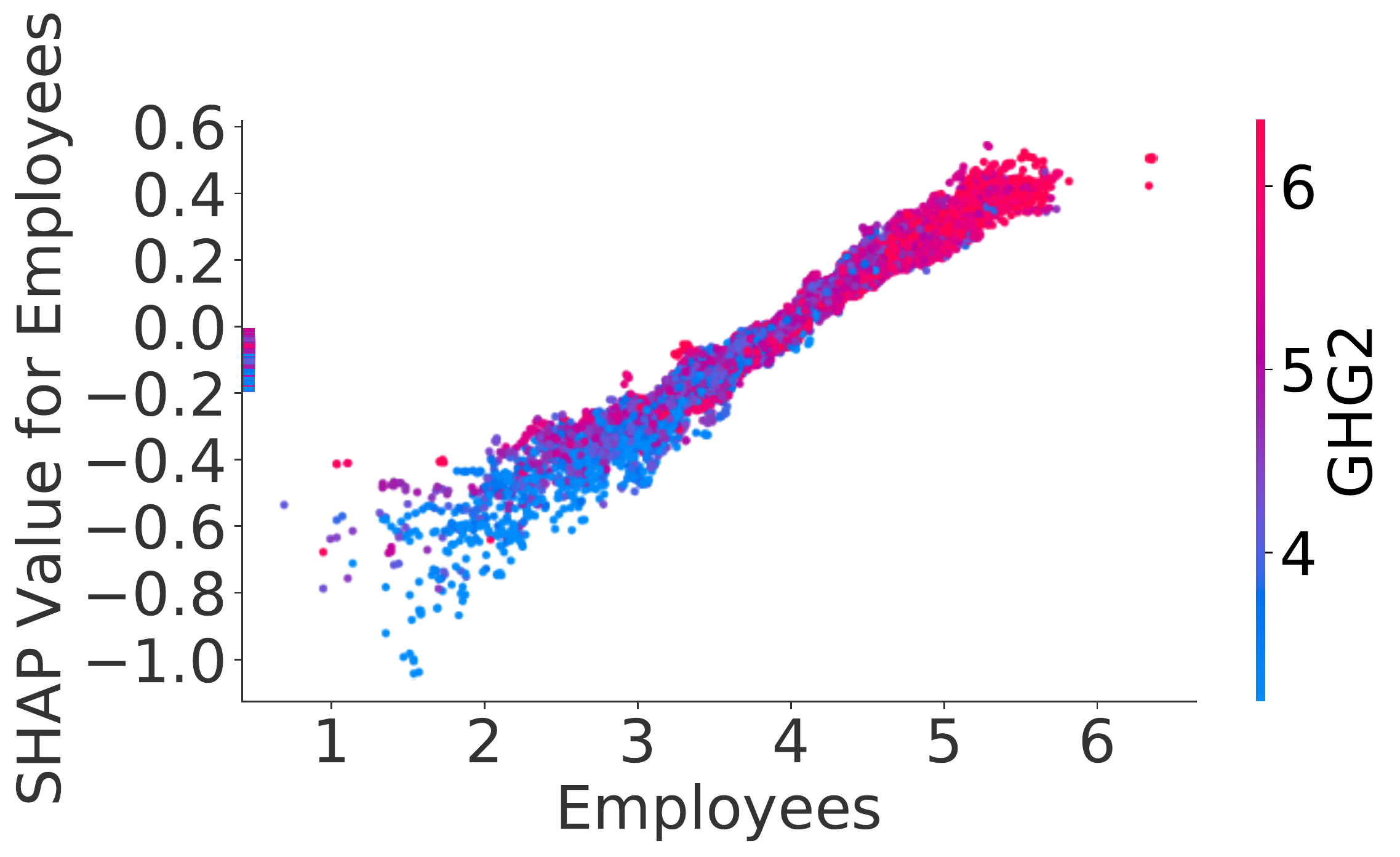}
		\caption{Scope 2}
		\label{fig:shap_CF2_Employees}
	\end{subfigure}
	
	\caption{Relationship between SHAP values of the Employees feature and the decimal logarithm of the Employees feature value.}
	\label{fig:shap_CF_Employees}
\end{figure}

Figure \ref{fig:shap_CF_Employees} shows a near-linear relationship between the SHAP values of the Employees feature and the decimal logarithm of the Employees feature values, apart from the few points on the Y-axis referring to missing data. 

\subsection{Categorical Data}
\label{app:shap_categorical_year}

\begin{figure}
	\centering
	\begin{subfigure}{0.49\textwidth}
		\centering
		\includegraphics[width=1\textwidth]{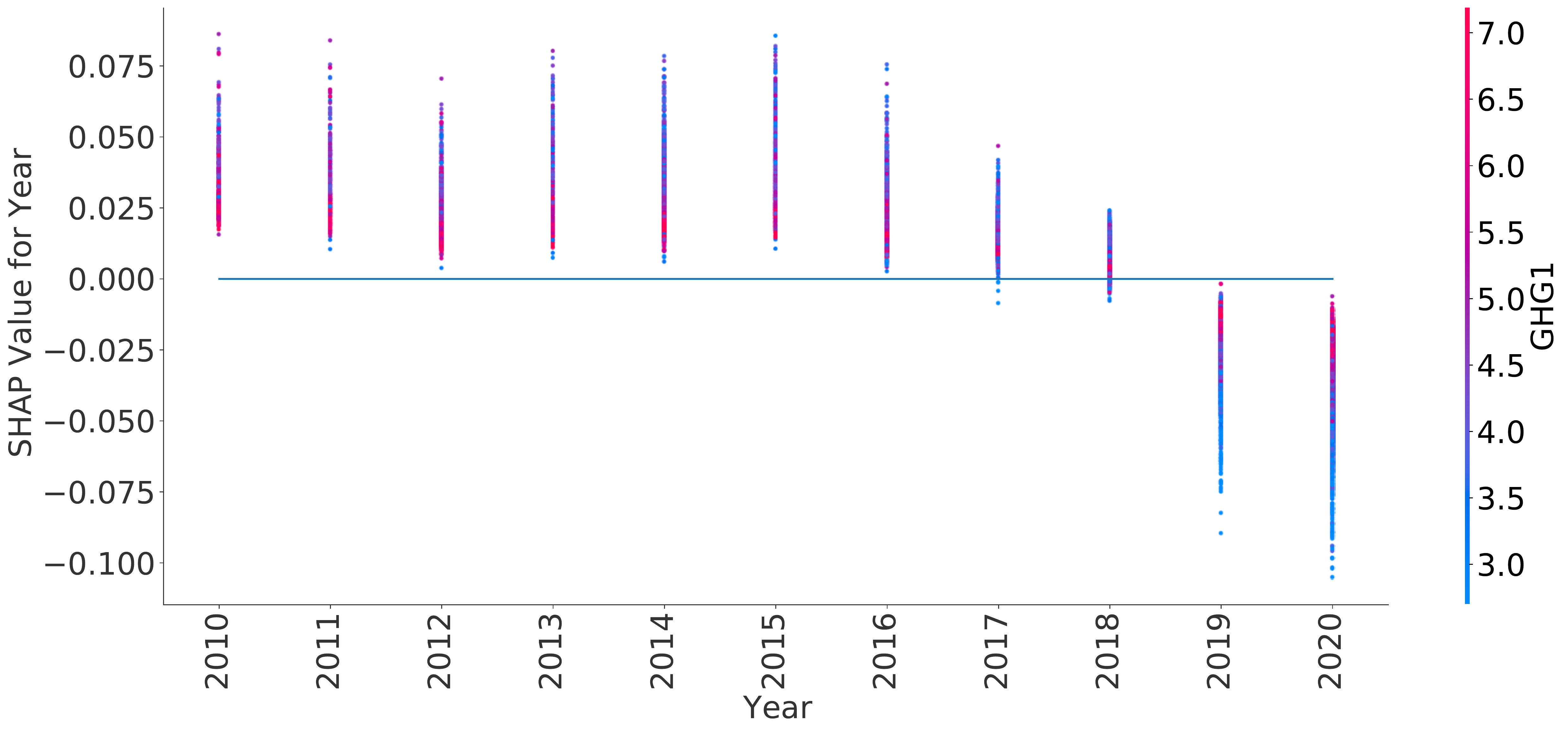}
		\caption{Scope 1}
		\label{fig:shap_CF1_Year}
	\end{subfigure}
	\hfill
	\begin{subfigure}{0.49\textwidth}
		\centering
		\includegraphics[width=1\textwidth]{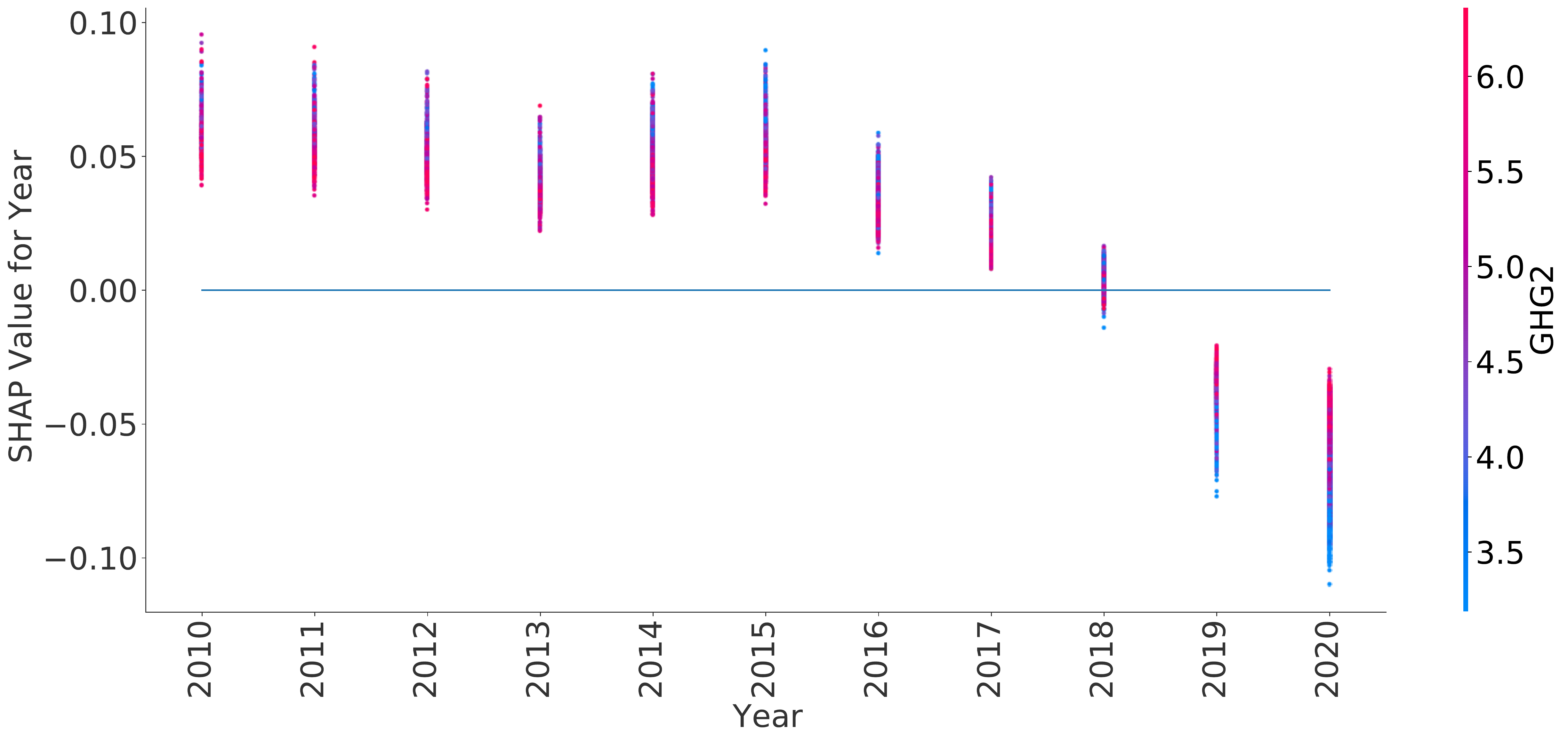}
		\caption{Scope 2}
		\label{fig:shap_CF2_Year}
	\end{subfigure}
	\caption{SHAP values: impact of the year on the predicted GHG emission.}
	\label{fig:shap_CF_Year}
\end{figure}

Figure \ref{fig:shap_CF_Year} shows the distribution of SHAP values for the Year feature, for each year in the training set. It is interesting to see that for both scope 1 and scope 2, the model captures a tendency to have lower GHG estimates as time passes.

\end{document}